\documentclass[times, review, 10pt]{elsarticle}

\usepackage{amsmath,amsfonts, amssymb}

\usepackage{array}
\usepackage{textcomp}
\usepackage{stfloats}
\usepackage{url}
\usepackage{verbatim}
\usepackage{color}   
\usepackage{booktabs}

\usepackage{graphicx}   

\usepackage[caption=false,font=normalsize,labelfont=normalsize,textfont=normalsize]{subfig}   

\usepackage{algorithmic}   
\usepackage{algorithm}

\usepackage{multirow}

\usepackage[T1]{fontenc}   

\newtheorem{definition}{Definition} 
\newtheorem{theorem}{Theorem}

\newtheorem{proposition}{Proposition}

\usepackage[capitalize,noabbrev]{cleveref}
\crefname{section}{Sec.}{Secs.}
\Crefname{section}{Section}{Sections}
\crefname{table}{Tab.}{Tabs.}
\Crefname{table}{Table}{Tables}
\crefname{figure}{Fig.}{Figs.}
\Crefname{figure}{Figure}{Figures}
\crefname{equation}{Eq.}{Eqs.}
\Crefname{equation}{Equation}{Equations}

\journal{Pattern Recognition}

\begin{document}
	
\begin{frontmatter}
    
    
    
    \title{Entropy-Informed Weighting Channel Normalizing Flow for Deep Generative Models}
    
    
    \author[1]{Wei Chen\corref{cor2}}
    \ead{maweichen@mail.scut.edu.cn}
    
    \author[2]{Shian Du\corref{cor2}}
    \ead{dsa23@mails.tsinghua.edu.cn}
    
    \author[1]{Shigui Li\corref{cor2}}
    \ead{lishigui@mail.scut.edu.cn}
    
    \author[3]{Delu Zeng\corref{cor1}}  
    \ead{dlzeng@scut.edu.cn}
    
    \author[4]{John Paisley}
    \ead{jpaisley@columbia.edu}
    
    \address[1]{School of Mathematics, South China University of Technology, Guangzhou, China}
    \address[2]{Shenzhen International Graduate School, Tsinghua University, Shenzhen, China}
    \address[3]{School of Electronic and Information Engineering, South China University of Technology, Guangzhou, China}
    \address[4]{Department of Electrical Engineering, Columbia
        University, New York, NY USA}
    
    \cortext[cor1]{Corresponding author}
    \cortext[cor2]{Joint contribution}
    
    

\begin{abstract}
    Normalizing Flows (NFs) are widely used in deep generative models for their exact likelihood estimation and efficient sampling. 
    However, they require substantial memory since the latent space matches the input dimension.
    Multi-scale architectures address this by progressively reducing latent dimensions while preserving reversibility.
    Existing multi-scale architectures use simple, static channel-wise splitting, limiting expressiveness. To improve this, we introduce a regularized, feature-dependent  $\mathtt{Shuffle}$ operation  and integrate it into vanilla multi-scale architecture. 
    This operation adaptively generates channel-wise weights and shuffles latent variables before splitting them.
    We observe that such operation guides the variables to evolve in the direction of entropy increase, hence we refer to NFs with the $\mathtt{Shuffle}$ operation as  \emph{Entropy-Informed Weighting Channel Normalizing Flow} (EIW-Flow). 
    Extensive experiments on CIFAR-10, CelebA, ImageNet, and LSUN demonstrate that EIW-Flow achieves state-of-the-art density estimation and competitive sample quality for deep generative modeling, with minimal computational overhead.
    
\end{abstract}


    
    
    
    
    
    
\end{frontmatter}



	
\section{Introduction}
Deep generative models are a type of machine learning paradigm that revolves around learning the probability distribution of data. These models are designed to generate new samples based on the learned data distribution, especially when there is a lack of data available for training. Although learning the true data distribution can be challenging due to practical constraints, deep generative models have been shown to approximate it effectively \cite{salakhutdinov2015learning}. Consequently, they have become highly valuable for a wide range of downstream pattern recognition tasks, such as image synthesis \cite{wan2026unified}, low-light image enhancement \cite{xu2025upt}, sarcasm detection \cite{zhang2026sarcasm}, anomaly detection \cite{hu2025msattnflow}. In recent years, deep generative models have also attracted considerable attention in emerging application domains, such as physiological signal modeling \cite{neifar2025deep} and materials discovery \cite{hellman2025brief}, leading to significant advances in the field.

Popular deep generative models like Generative Adversarial Networks (GANs) \cite{aggarwal2021generative} model the underlying true data distribution implicitly, but do not employ the maximum likelihood criterion for analysis. Variational autoencoders (VAEs) \cite{zhang2018advances} map data into a low-dimensional latent space for efficient training, resulting in optimizing a lower bound of the log-likelihood of data. Different from GANs and VAEs, Autoregressive Flows \cite{kingma2016improved} and Normalizing Flows are both inference models that directly learn the log-likelihood of data. Although Autoregressive Flows excel at capturing complex and long-term dependencies within data dimensions, their sequential synthesis process makes parallel processing difficult, leading to limited sampling speed.

Normalizing Flows (NFs) define a series of reversible transformations between the true data distribution and a known base distribution (e.g.,  a standard normal distribution). These transformations are elucidated in works like Real NVP \cite{dinh2016density} and GLOW \cite{kingma2018glow}. NFs have the advantage of exact estimation for the likelihood of data. Additionally, the sampling process of NFs is considerably easier to parallelize than Autoregressive Flows and can be optimized directly compared to GANs. However, a significant challenge for NFs lies in efficiently achieving a high level of expressive power in the high-dimensional latent space. This challenge arises because data typically lies in a high-dimensional manifold, and it is essential for NFs to maintain the dimension across the inference or sampling process to ensure its reversibility. 
	
In order to mitigate this problem, Real NVP introduced a multi-scale architecture that progressively reduces the dimension of latent variables while ensuring its reversibility. This architecture aids in accelerating training and save memory resources.
However, the $\mathtt{Split}$ operation employed in the multi-scale architecture compromises NFs' expressive power as it simply splits the high-dimensional variable into two lower-dimensional variables of equal size. As a result, it does not take into account the interdependencies between different channel feature maps of the variable, which can restrict the NFs' overall expressive power. 
	
\begin{figure*}[htb]
    \centering
    \includegraphics[width=0.99\linewidth]{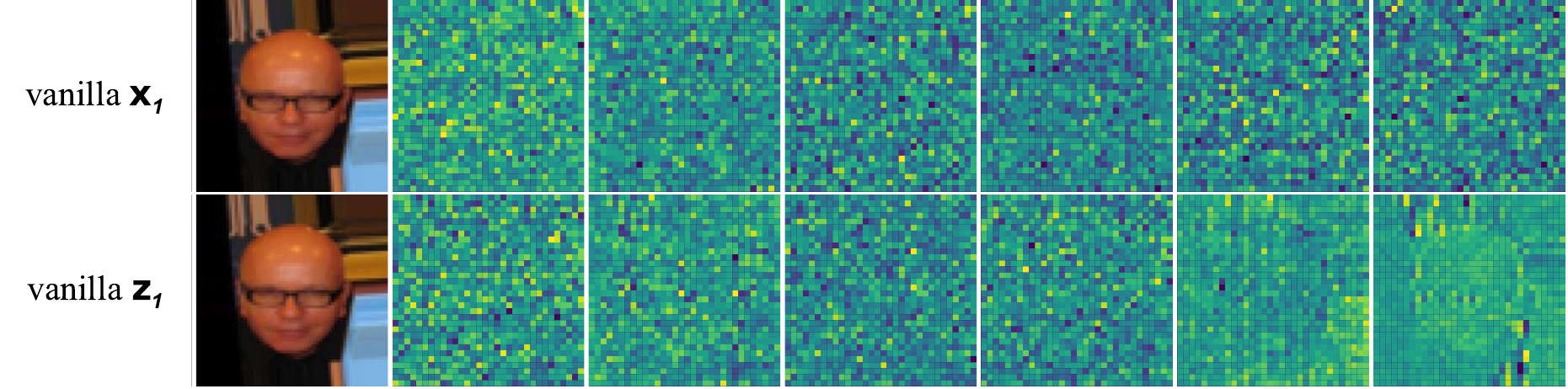}  %
    \caption{Channel feature maps before the $\mathtt{Split}$ operation (without applying our proposed $\mathtt{Shuffle}$ operation) in a multi-scale architecture. The first column shows  CelebA faces, and the other columns show their corresponding feature maps. The top row displays the feature maps of  $\mathbf{x}_1$ and the bottom row those of $\mathbf{z}_1$, both from the first scale of the NFs.}    
    \label{fig:celeba-middle-status-train-vanilla}
\end{figure*}

As illustrated in \cref{fig:celeba-middle-status-train-vanilla}, some channel feature maps of $\mathbf{z}_1$ capture  facial contour information. However, since   $\mathbf{z}_1$ is assumed to follow a Gaussian distribution in NFs, traditional multi-scale NFs fail to retain such information, limiting their effectiveness and expressiveness.

\begin{figure*}[ht]
    \centering
    \includegraphics[width=0.99\linewidth]{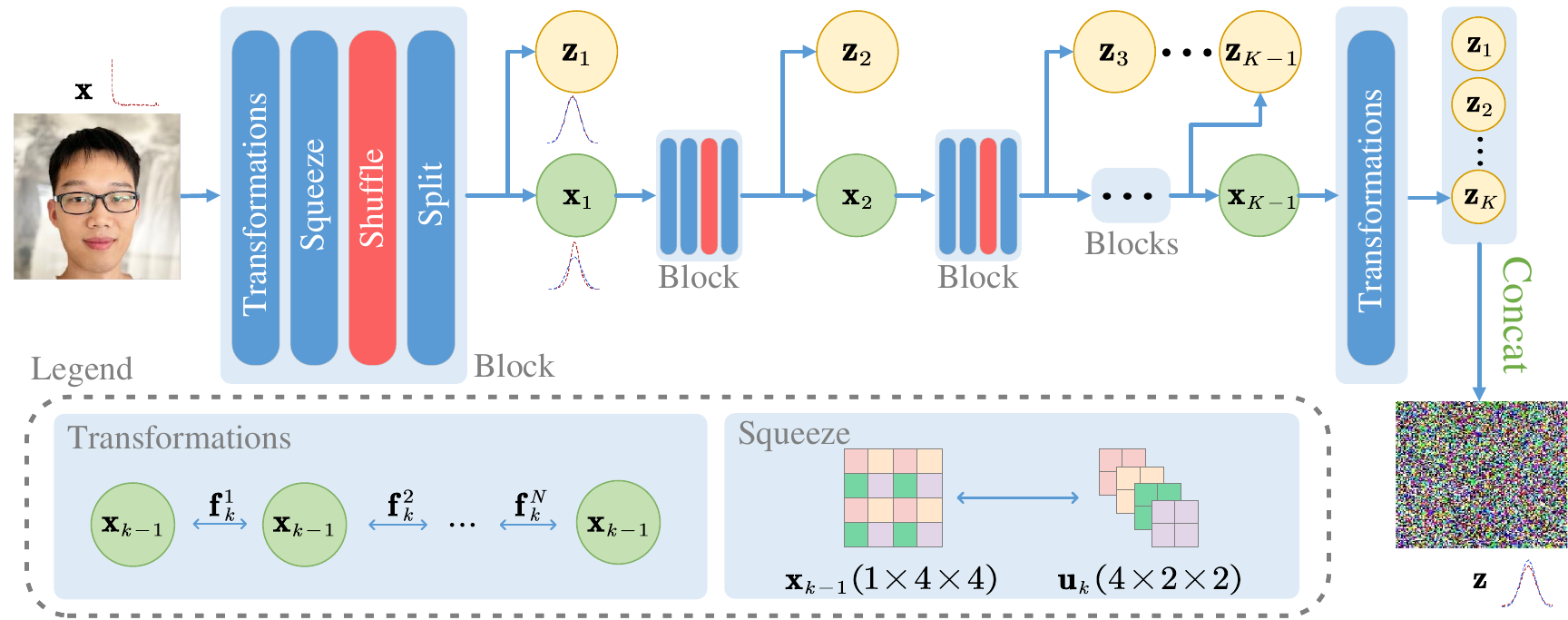}
    \caption{Schematic diagram of the  inference process of EIW-Flow with multi-scale architecture. Compared to vanilla architecture, an added  $\mathtt{Shuffle}$ operation adaptively propagates  information-rich channel feature across scales. Red and blue dashed lines indicate the true distribution and standard Gaussian distribution of a latent variable, respectively.}
    \label{fig:diagram}
\end{figure*}

To address this, we propose a regularized $\mathtt{Shuffle}$ operation  before $\mathtt{Split}$ operation in the inference process. This operation assigns weights to channel feature maps based on their information content, then shuffling them accordingly while preserving reversibility. More importantly, we demonstrate that channel feature maps with higher entropy, which are more likely to follow a Gaussian distribution, are prioritized for forming the final latent variable. Based on this, we enhance  the conventional multi-scale architecture and propose a novel normalizing flow model called "\textbf{E}ntropy-\textbf{I}nformed \textbf{W}eighting Channel Normalizing \textbf{Flow}" (EIW-Flow), with its schematic diagram shown in \cref{fig:diagram}.

EIW-Flow's key innovation is its entropy-informed adaptive splitting mechanism. Unlike traditional multi-scale architectures that split latent variables equally, discarding feature-rich channels prematurely (\cref{fig:celeba-middle-status-train-vanilla}), EIW-Flow dynamically reorders channels via entropy-guided $\mathtt{Shuffle}$ (\cref{sec:method}). This ensures $\mathbf{z}_1$ aligns with the Gaussian assumption while preserving structural information in $\mathbf{x}_1$.

The main contributions in this paper are as follows: 
\begin{enumerate}
    \item We design a regularized $\mathtt{Shuffle}$ operation to adaptively shuffle the channel feature maps of its input based on the feature information contained in each map, while maintaining the reversibility of NFs.
    
    \item We propose to divide the $\mathtt{Shuffle}$ operation into three distinct components: the \emph{solver}-$\mathcal{S}$, the  \emph{guider}-$\mathcal{G}$ and the \emph{shuffler}-$\mathcal{S}_{\mathcal{F}}$ (see \cref{sec:necessity} for details). 
		
    \item We demonstrate the efficacy of the $\mathtt{Shuffle}$ operation from the perspective of entropy using the principles of information theory and statistics, such as the Central Limit Theorem and the Maximum Entropy Principle. 
    
    \item Our experiments show state-of-the-art results on density estimation and comparable sampling quality with a negligible additional computational overhead.
\end{enumerate}

\section{Related Works}
\begin{figure*}[!ht]
    \begin{center}
        \includegraphics[width=\linewidth]{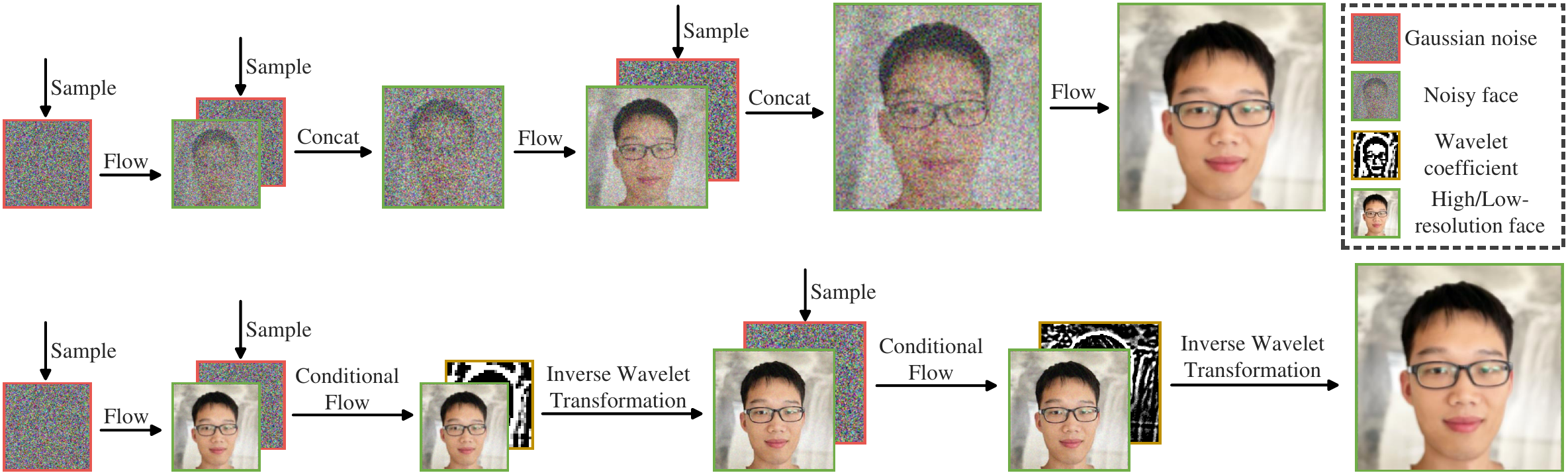}
    \end{center}
    \caption{The sampling process of the multi-scale architectures in EIW-Flow (top) and Wavelet Flow (bottom). "Conditional flow" and "inverse wavelet transformation" are introduced in \cite{yu2020wavelet}. "Sample" refers to drawing Gaussian noise from $\mathcal{N}(\mathbf{0},\mathbf{I})$, while "Flow" denotes NF steps.
    }
    \label{fig:Waveletflow}
\end{figure*}

\textbf{Multi-scale architectures} have been extensively investigated in deep generative models for latent space dimension reduction. Some initial works \cite{dinh2016density} suggested implementing a $\mathtt{Split}$ operation, which splits latent variables at earlier scales to distribute the loss function throughout flow models, leading to a substantial reduction in computation and memory. \cite{kingma2018glow} introduced  unsqueeze and  squeeze operations before and after the the $\mathtt{Split}$ operation. 
However, the $\mathtt{Split}$ operation employed in the multi-scale architecture compromises the expressive power of NFs as it simply splits the high-dimensional variable into two lower-dimensional variables of equal size without considering the feature information of its input. 
\cite{yu2020wavelet} proposed to replace the $\mathtt{Split}$ operation with the wavelet transformation, which splits high-resolution variables into low-resolution variables and corresponding wavelet coefficients. This paper proposes to consider feature information contained in each channel feature map of high-dimensional variable during the $\mathtt{Split}$ operation. 
The schematic diagrams of the sampling process of our model and Wavelet Flow are shown in  \cref{fig:Waveletflow}.
In the sampling process, Wavelet Flow progressively transforms an image from low-resolution to high-resolution by predicting corresponding wavelet coefficients and applying wavelet transformation. Nevertheless, EIW-Flow (ours) progressively concatenates low-dimensional noisy variable with Gaussian noise to get high-dimensional variable and removes Gaussian noise from noisy variable with flow steps.

\textbf{Image Captioning.} 
Recent advances in image captioning further highlight the importance of hierarchical feature representation and dynamic architectures, which are conceptually aligned with our work on normalizing flows. 
For instance, LSTNet \citep{ma2023towards} preserves spatial relationships through locality-sensitive attention over grid features, which shares conceptual similarities with our adaptive shuffle mechanism. 
Similarly, CAST \citep{cao2024cast} introduces cross-modal retrieval and visual conditioning to improve caption quality.
Our EIW-Flow achieves similar spatial consistency by performing entropy-aware channel splitting before latent partitioning, ensuring that high-entropy features are retained in $\mathbf{x}_k$. 
The DTNet \citep{ma2024image} complements these ideas by customizing network paths based on input characteristics, conceptually resonating with our entropy-increasing principle and adaptive feature flow. 
Taken together, these works emphasize three insights: (1) preserving fine-grained spatial relationships is critical for high-quality generation, (2) dynamic architectural adaptation is an emerging paradigm for improving expressiveness, and (3) multi-scale feature integration is essential for capturing hierarchical semantics. These theoretical underpinnings support our entropy-guided design in normalizing flows.
	
\section{Background}
In this section, we review the basics of deep generative models and Normalizing Flows. We then discuss the multi-scale architecture and its bottleneck problem.

\subsection{Deep Generative Models}
Let $\mathbf{x}$ be a random variable with complex unknown distribution $p(\mathbf{x})$ and i.i.d. dataset $\mathcal{X}$. The main goal of generative models is to define a parametric distribution $q_{\boldsymbol{\theta}}(\mathbf{x})$ and learn its optimal parameters $\boldsymbol{\theta}^{*}$ to approximate the true but unknown data distribution $p(\mathbf{x})$ \cite{du2022flow}.
An efficient and tractable criterion is Maximum Likelihood (ML):
\begin{equation}
    \label{ml}
    \boldsymbol{\theta}^{*}=\mathop{\arg\max}\limits_{\boldsymbol{\theta} \in \boldsymbol{\Theta}}\mathbb{E}_{\mathbf{x}\sim p(\mathbf{x})}[\log q_{\boldsymbol{\theta}}(\mathbf{x})].
\end{equation}

\subsection{Normalizing Flows}
Normalizing Flows (NFs) define an isomorphism $\mathbf{f}$ with parameters $\boldsymbol{\theta}$ that  transform the observed data $\mathbf{x}$ in $\mathcal{X}$ to a corresponding latent variable $\mathbf{z}$ in a latent space $\mathcal{Z}$,
\begin{equation}
    \mathbf{z} = \mathbf{f}(\mathbf{x}) \quad \text{and} \quad \mathbf{x} = \mathbf{f}^{-1}(\mathbf{z}). 
\end{equation}
	
In practice, $\mathbf{f}$ is usually composed of a series of isomorphisms $\{ \mathbf{f}_{k} \}_{k=1}^{K}$, i.e. $\mathbf{f}=\mathbf{f}_{K}\circ \mathbf{f}_{K-1}\circ \cdots \circ \mathbf{f}_{1}$. And each $\mathbf{f}_k$   transforms intermediate variable $\mathbf{x}_{k-1}$ to $\mathbf{x}_{k}$, where  $\mathbf{x}_{0}\triangleq\mathbf{x}$ and $\mathbf{x}_{K}\triangleq \mathbf{z}$. $\mathbf{z}$ follows a tractable and simple distribution such as isotropic Gaussian distribution $N(\boldsymbol{\mu}, \mathbf{I})$. Consequently,  NFs can be summarized as:
\begin{equation}
    \label{transform}
    \mathbf{z}\stackrel{\mathbf{f}_{K}}{\longleftrightarrow} \cdots \stackrel{\mathbf{f}_{3}}{\longleftrightarrow}\mathbf{x}_{2}\stackrel{\mathbf{f}_{2}}{\longleftrightarrow} \mathbf{x}_{1}\stackrel{\mathbf{f}_{1}}{\longleftrightarrow} \mathbf{x}.
\end{equation}
	
Given that we can compute the log-likelihood $q_{\mathbf{z}}(\mathbf{z})$ of $\mathbf{z}$, the unknown likelihood $q_{\boldsymbol{\theta}}(\mathbf{x})$ of $\mathbf{x}$ under the transformations $\{ \mathbf{f}_{k} \}_{k=1}^{K}$ can be computed by using the change of variables formula,
\begin{equation}
    \label{change}
    \log q_{\mathbf{\boldsymbol{\theta}}}(\mathbf{x})= \log q_{\mathbf{z}}(\mathbf{z})+\log\left| \det \left(  \frac{\partial \mathbf{f}_{K}  \circ \cdots \circ  \mathbf{f}_{1} (\mathbf{x} )}{\partial \mathbf{x}}  \right)\right|=\log q_{\mathbf{z}}(\mathbf{z})+\sum_{k=1}^{K}\log\left| \det(\mathbf{J}_{k})\right|,
\end{equation}
where $\mathbf{J}_{k}=\partial \mathbf{f}_{k}(\mathbf{x}_{k-1})/\partial \mathbf{x}_{k-1}$ denotes the Jacobian of isomorphism $\mathbf{f}_{k}$ and any two Jacobians follow the fact that $\det(A \cdot B)=\det(A)\cdot \det(B)$.
	
NFs allow for a uniquely reversible encoding and exact likelihood computation. They are much more stable than other deep generative models like GANs. However, there still exist some limitations which hamper their expressive power. First, to ensure the reversibility of the model, the class of transformations $\{\mathbf{f}_{k}\}_{k=1}^{K}$ is constrained and therefore sacrifices their expressive power \cite{bhattacharyya2020normalizing}. Second, the dimension of the variables must remain the same during the transformation in order to ensure that the log-det term in  \cref{change} can be computed. For high-dimensional image data, unnaturally forcing the width of neural network to be the same as the data dimension greatly hinders further development of NFs and also reduces the flexibility to a certain extent.

In order to approximate the true data distribution $p(\mathbf{x})$, NFs are usually stacked very deep, which requires significantly more training time than other generative models. To speed up training and save the memory, a simple yet effective architecture named Multi-Scale Architecture was proposed in \cite{dinh2016density} and can be combined with most NFs.

\subsection{Multi-Scale Architecture}
\label{multiscale}
Multi-scale architecture contains a number of scales, each with different spatial and channel sizes for the variable propagating through it. 

Assuming our model consists of $K$ scales and each scale has $N$ steps of transformation. At $k$-th scale ($k=1,2,\ldots,K-1$), there are several operations combined into a sequence. Firstly, several transformations $\{ \mathbf{f}^{i}_{k}\}_{i=1}^{N}$ with $\mathbf{f}_k=\mathbf{f}_{k}^{N}\circ\cdots \circ \mathbf{f}_{k}^{2}\circ\mathbf{f}_{k}^{1}$ are applied  $\mathbf{x}_{k-1}$  while keeping its shape $[C_k, H_k, W_k]$, where $C_k$ and $[H_k, W_k]$ denote the channel and spatial size at $k$-th scale, respectively (see \cref{i-transform}). 

Then a $\mathtt{Squeeze}$ operation is performed to reshape the variable $\mathbf{x}_{k-1} $ with shape $[C_k, H_k, W_k]$ to $\mathbf{u}_k $ with shape $[4C_k, H_k/2, W_k/2]$ by reshaping $2\times2$ neighborhoods into $4$ channels\cite{dinh2016density}, which effectively trades spatial size for numbers of channels. Unless otherwise specified, $\mathbf{u}_k $ is the variable after $\mathtt{Squeeze}$ in the inference process.

Finally,  the variable $\mathbf{u}_k $ with shape $[4C_k, H_k/2, W_k/2]$ is split into two equal parts $\mathbf{x}_k$ and $ \mathbf{z}_k$ at channel level, both of which share the same shape $[2C_k, H_k/2, W_k/2]$. Moreover, $\mathbf{x}_k$ is propagated to the next scale so that its more abstract spatial and channel features can be further extracted, while the other half $\mathbf{z}_k$ is left unchanged to form $\mathbf{z}$. 

The complete process at the $k$-th scale is shown as follows:
\begin{align}
    \mathbf{u} _k &= \mathtt{Squeeze}(\mathbf{f}_{k}^{N}\circ\cdots \circ\mathbf{f}_{k}^{2}\circ\mathbf{f}_{k}^{1}(\mathbf{x}_{k-1})),\label{i-transform}\\
    [\mathbf{x}_k, \mathbf{z}_k] &= \mathtt{Split}(\mathbf{u} _k)  \label{split}.
\end{align}

Note that for the final scale, the $\mathtt{Squeeze}$ and $\mathtt{Split}$ operations are not performed on $\mathbf{u}_{K}$, hence the output of the final scale is $\mathbf{z}_{K}=\mathbf{f}_{K}^{N}\circ\cdots \circ\mathbf{f}_{K}^{2}\circ\mathbf{f}_{K}^{1}(\mathbf{x}_{K-1})$. 

After all scales, the unchanged half $\{ \mathbf{z}_k\}_{k=1}^{K}$ at each scale are collected and reshaped appropriately, and then concatenated at channel level to form the final latent variable $\mathbf{z}$:
\begin{equation}
    \label{concat}
    \mathbf{z}=\mathtt{Concat}(\mathbf{z}_1, \mathbf{z}_2,..., \mathbf{z}_K).
\end{equation}

A key drawback of traditional multi-scale architectures is the static splitting of latent variables (\cref{split}). Dividing $ \mathbf{u}_k $ into equal halves $[\mathbf{x}_k, \mathbf{z}_k]$ may assign structural features to $ \mathbf{z}_k $, violating the Gaussian assumption. This mismatch reduces expressiveness, as $ \mathbf{z}_k$ deviates from a simple distribution and $ \mathbf{x}_k $ loses important details. EIW-Flow addresses this by adaptively shuffling channels before splitting, ensuring $ \mathbf{z}_k $ retains high-entropy channels.

\section{Method}
\label{sec:method}
\begin{figure}[!ht]
    \begin{center}
        \includegraphics[width=\linewidth]{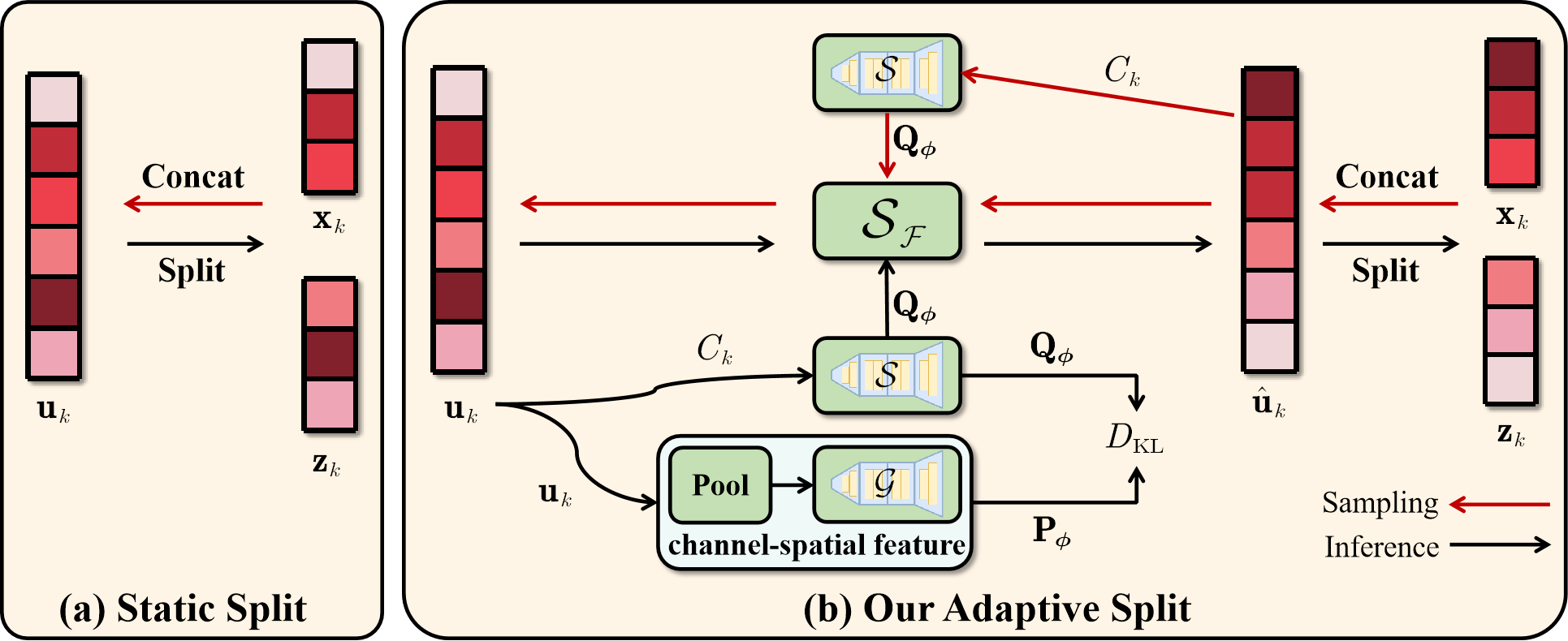}
    \end{center}
    \caption{Schematic diagram of the inference and sampling process of the $\mathtt{Shuffle}$ and $\mathtt{Split}$ operations. This figure compares two methods of $\mathtt{Split}$ operation: (a) the  static $\mathtt{Split}$ used in traditional NFs, and (b) our proposed adaptive $\mathtt{Split}$ operation. $\mathcal{S}$, $\mathcal{G}$ and $\mathcal{S}_{\mathcal{F}}$ are our designed \emph{solver} , \emph{guider} and \emph{shuffler}, respectively.  For the channel feature maps of $\mathbf{u}_k$ and $\hat{\mathbf{u}}_k$, the darker the color, the greater the corresponding element in $\mathbf{Q}_{\boldsymbol{\phi}}$. 
    }
    \label{fig:diagramofselect}
\end{figure}

In this section, a regularized $\mathtt{Shuffle}$ operation is introduced, which adaptively shuffles channel feature maps of latent variables. 
Information-rich features are propagated to the next scale, while others form the final latent variable $\mathbf{z} \sim \mathcal{N}(\mathbf{0}, \mathbf{I})$. The $\mathtt{Shuffle}$ operation consists of three components: a \emph{solver}-$\mathcal{S}$ (see \cref{solver}), a  \emph{guider}-$\mathcal{G}$ (see \cref{guider}) and a \emph{shuffler}-$\mathcal{S}_{\mathcal{F}}$ (see \cref{select}). Besides, a penalty term is added to the objective function (see \cref{sec:objective}) to regularize  $\mathtt{Shuffle}$.  The schematic diagram of  $\mathtt{Shuffle}$ is shown in \cref{fig:diagramofselect}.
	
\subsection{Necessity of Solver and Guider}
\label{sec:necessity}
	
To propagate information-rich channel features across scales, $\mathtt{Shuffle}$ must balance reversibility and expressiveness, which is difficult with a single network (see \cref{solver}). 
To address this, we introduce two separate networks. 
One network, called \emph{solver}-$\mathcal{S}$, is reversible and assigns weights to all channel feature maps without absorbing feature information. The other, called \emph{guider}-$\mathcal{G}$, used only in the inference process, extracts feature information from latent variables. Acting as a teacher, \emph{guider}-$\mathcal{G}$ guides \emph{solver}-$\mathcal{S}$ to optimize the $\mathtt{Shuffle}$ operation efficiently. 
	
\subsection{Entropy-informed Channel Weight Solver}
\label{solver}
	
The \emph{solver}-$\mathcal{S}$ generates and assigns channel-wise weights to the channel feature maps of latent variables. 
In the inference process, weights are assigned to $\{\mathbf{u}_k\}_{k=1}^{K-1}$ before the $\mathtt{Split}$ operation, whereas in sampling, weights are assigned to  $\{\hat{\mathbf{u}}_k\}_{k=1}^{K-1}$ after the $\mathtt{Concat}$ operation.
This assignment is based on the richness and importance of feature information in each channel feature map of $\{\mathbf{u}_k\}_{k=1}^{K-1}$ or $\{\hat{\mathbf{u}}_k\}_{k=1}^{K-1}$, quantified by entropy according to the principles of information theory (\cref{sec:analysis}). 
Initially, $\mathcal{S}$ assigns equal weights to all channel feature maps. 
	
A key challenge arises: if $\mathcal{S}$ directly takes $\{\mathbf{u}_k\}_{k=1}^{K-1}$ as input in the inference process, it cannot be reverted in the sampling process because the $\mathtt{Concat}$ operation precedes  $\mathtt{Shuffle}$ in sampling, shifting input from $\{\mathbf{u}_k\}_{k=1}^{K-1}$ to $\{\hat{\mathbf{u}}_k\}_{k=1}^{K-1}$ (see  \cref{fig:diagramofselect}).
To ensure reversibility, we use the channel number of $\mathbf{u}_{k}$, i.e., $4C_{k}$,  as input, making $\mathcal{S}$ inherently reversible at any scale. Here, reversibility means that $\mathcal{S}$ produces the same output in both inference and sampling. 

For the $k$-th scale in inference, $\mathcal{S}$ takes $\mathbf{u}_k$ as input, specifically its channel number $4C_k$. We denote the input as $\mathbf{u}_k$ for simplicity, understanding that only the channel count is used. 
The  structure of $\mathcal{S}$ is defined as: 
\begin{equation}
    \mathbf{Q}_{\boldsymbol{\phi},\mathbf{u}_k} = \mathcal{S}(\mathbf{u}_k, \boldsymbol{W}_{\mathcal{S}})=\mathtt{Softmax}(\boldsymbol{W}_\mathcal{S}^{L_{\mathcal{S}}} \mathbf{h}_{L_{\mathcal{S}}-1}(\cdots \mathbf{h}_{1}(\boldsymbol{W}_\mathcal{S}^{1} \mathbf{u}_k))), \label{generator}
\end{equation}
where $\mathbf{Q}_{\boldsymbol{\phi},\mathbf{u}_k}$ quantifies feature importance of $\mathbf{u}_k$, $\boldsymbol{\phi}$ are trainable parameters of the $\mathtt{Shuffle}$ operation. $\mathbf{Q}_{\boldsymbol{\phi},\mathbf{u}_k}$ is referred to as $\mathbf{Q}_{\boldsymbol{\phi}}$ for the sake of brevity throughout the paper. $\boldsymbol{W}_{\mathcal{S}}=\{\boldsymbol{W}_{\mathcal{S}}^1, \boldsymbol{W}_{\mathcal{S}}^2, \ldots, \boldsymbol{W}_{\mathcal{S}}^{L_{\mathcal{S}}}\}$ are  parameters of $L_{\mathcal{S}}$ layers in $\mathcal{S}$. 
Here $\boldsymbol{W}_\mathcal{S}^{1} \in \mathbb{R}^{2C_{k}\times 1}$, $\boldsymbol{W}_\mathcal{S}^{l} \in \mathbb{R}^{2C_{k}\times 2C_{k}}$ ($l=2,3,\ldots,L_{\mathcal{S}}-1$) and $\boldsymbol{W}_\mathcal{S}^{L_{\mathcal{S}}} \in \mathbb{R}^{4C_{k} \times 2C_{k}}$. $\{\mathbf{h}_{l}\}_{l=1}^{L_{\mathcal{S}}-1}$ are activation functions. 
The length of $\mathbf{Q}_{\boldsymbol{\phi}, \mathbf{u}_k}$ and $\mathbf{Q}_{\boldsymbol{\phi}, \hat{\mathbf{u}}_k}$ are $4C_{k}$, matching the channel number of $\mathbf{u}_k$ and $\hat{\mathbf{u}}_k$. 
In inference, $\hat{\mathbf{u}}_k$ is then split into two equal parts, $\mathbf{x}_k$ and  $\mathbf{z}_k$.
	
A major challenge is ensuring $\mathcal{S}$ correctly assigns weights, so that elements of elements in $\mathbf{Q}_{\boldsymbol{\phi}, \mathbf{u}_k}$ and $\mathbf{Q}_{\boldsymbol{\phi}, \hat{\mathbf{u}}_k}$ corresponding to $\mathbf{x}_k$ are greater than those for $\mathbf{z}_k$, indicating that $\mathbf{x}_k$ contains more important channel feature maps.
However, since $\emph{solver}$-$\mathcal{S}$ only takes the channel number as input, it lacks direct access to feature information. 
Inspired by knowledge distillation\cite{wang2021joint}, we introduce a feature extraction network as a teacher to guide $\mathcal{S}$ in learning  feature information, termed the  \textbf{\textit{Supervisory Weight Guider}}.
	
\subsection{Supervisory Weight Guider}
\label{guider}
	
The \emph{guider}-$\mathcal{G}$ generates channel-wise weights by extracting feature information from latent variables. Unlike $\mathcal{S}$, it is irreversible and used only in the inference process. 
To construct $\mathcal{G}$, we refine global information by compressing spatial features, converting each 2D channel feature map into a scalar via pooling technique \cite{hsiao2019filter}. 
This method enhances robustness to spatial transformations while adding negligible trainable parameters, reducing overfitting risk.
	
Formally, at scale $k$, $\mathcal{G}$ takes input $\mathbf{u}_k = [\mathbf{u}_k^1, \mathbf{u}_k^2, \ldots , \mathbf{u}_k^{4C_{k}} ]^{T}\in \mathbb{R}^{4C_k \times H_k/2 \times W_k/2}$, where $\mathbf{u}_k^c\in \mathbb{R}^{\frac{H_k}{2} \times \frac{W_k}{2}}$ is the $c$-th channel feature map. 
The global average pooling compress for each channel feature map can be described as:
	\begin{equation}
		\bar{u}_{k}^{c}=\mathtt{Pool}(\mathbf{u}_k^{c})=\frac{4}{H_k \times W_k} \sum_{a=1}^{H_k/2} \sum_{b=1}^{W_k/2} \mathbf{u}_k^{c}(a, b),   \label{squeezeSENet}
	\end{equation}
	where $\mathtt{Pool}$ generates channel-wise feature $\bar{u}_{k}^{c}$. For ease of notation,  $\mathbf{u}_k^{c}(a, b)$ is the value at $(a, b)$ of $\mathbf{u}_k^{c}$, $\bar{u}_{k}^{c} \in \mathbb{R}^{1 \times 1}$ contains the global feature information of $\mathbf{u}_k^{c}$ and $\bar{\mathbf{u}}_{k}=[\bar{u}_{k}^{1}, \bar{u}_{k}^{2}, \ldots, \bar{u}_{k}^{4C_k}]^{T} \in \mathbb{R}^{4C_k\times 1\times 1}$ is the output of $\mathtt{Pool}(\mathbf{u}_k)$.
	
Next, $\mathcal{G}$ processes these pooled features to generate the weight vector $\mathbf{P}_{\boldsymbol{\phi},\mathbf{u}_{k}}$, representing feature importance:
\begin{equation}
        \mathbf{P}_{\boldsymbol{\phi}, \mathbf{u}_{k}}=\mathcal{G}\left(\mathbf{u}_{k}, \mathbf{W}_{\mathcal{G}}\right) 
        =\mathtt{Softmax}(\mathbf{W}_{\mathcal{G}}^{L_{\mathcal{G}}} \mathbf{h}_{L_{\mathcal{G}}-1}(\cdots \mathbf{h}_{1}(\mathbf{W}_{\mathcal{G}}^{1} \mathtt{Pool}(\mathbf{u}_{k}) ))),
\end{equation}
where $\mathbf{P}_{\boldsymbol{\phi},\mathbf{u}_k}$ denotes the importance and richness quantification vector outputted by $\mathcal{G}$ given the input $\mathbf{u}_k$. 
$\mathbf{P}_{\boldsymbol{\phi},\mathbf{u}_k}$ is referred to as $\mathbf{P}_{\boldsymbol{\phi}}$ for simplicity throughout the paper. $\boldsymbol{W}_{\mathcal{G}}=\{\boldsymbol{W}_{\mathcal{G}}^1, \boldsymbol{W}_{\mathcal{G}}^2, \ldots, \boldsymbol{W}_{\mathcal{G}}^{L_{\mathcal{G}}}\}$ are parameters of $L_{\mathcal{G}}$ layers in $\mathcal{G}$. Here $\boldsymbol{W}_\mathcal{G}^{1} \in \mathbb{R}^{\lfloor\frac{4C_{k}}{r}\rfloor \times 4C_k}$, $\boldsymbol{W}_\mathcal{G}^{l} \in \mathbb{R}^{\lfloor\frac{4C_{k}}{r}\rfloor \times \lfloor\frac{4C_{k}}{r}\rfloor} (l=2,3,\ldots,L_{\mathcal{G}}-1)$ and $\boldsymbol{W}_\mathcal{G}^{L_{\mathcal{G}}} \in \mathbb{R}^{4C_{k} \times \lfloor\frac{4C_{k}}{r}\rfloor}$. 
The reduction ratio $r$ improves efficiency, and $\{\mathbf{h}_{l}\}_{l=1}^{L_{\mathcal{G}}-1}$ are activation functions.
	
\subsection{Reversible and Entropy-informed Shuffle Operation}
\label{select}
After the \emph{solver}-$\mathcal{S}$ generates  channel-wise weights, the \emph{shuffler}-$\mathcal{S}_{\mathcal{F}}$ shuffles the latent variables, such as $\mathbf{u}_k$ or $\hat{\mathbf{u}}_k$, across channel dimensions based on these weights. In the following part, we will summarize the $\mathtt{Shuffle}$ operation at $k$-th scale used in both the inference and sampling process.
	
In inference, $\mathcal{S}$ takes the channel number of  $\mathbf{u}_k$, i.e., $C_k$, as input and outputs a $C_k$-dimensional vector $\mathbf{Q}_{\boldsymbol{\phi},\mathbf{u}_k}$. 
This guides the $\mathtt{Shuffle}$ operation, yielding $\hat{\mathbf{u}}_k$, which is then split into $\mathbf{x}_k$ and $\mathbf{z}_k$ along the channel dimension. 
	
In sampling, $\mathcal{S}$ operates inversely.
Given the channel number of $\hat{\mathbf{u}}_k$, i.e., $C_k$, it outputs the importance vector $\mathbf{Q}_{\boldsymbol{\phi}, \hat{\mathbf{u}}_k}$. 
Using $\hat{\mathbf{u}}_k$, $\mathcal{S}$ reconstructs the importance of each channel, aiding in recovering $\mathbf{u}k$. While $\mathtt{Shuffle}$ is not mathematically reversible, it enables information restoration via $\mathbf{Q}_{\boldsymbol{\phi}, \hat{\mathbf{u}}_k}$.
	
\subsection{Objective Function}
\label{sec:objective} 
The importance vector$\mathbf{P}_{\boldsymbol{\phi},\mathbf{u}_k}$ guides $\mathbf{Q}_{\boldsymbol{\phi},\mathbf{u}_k}$ in extracting feature information from $\mathbf{u}k$. 
Their components sum to 1, representing the distribution of channel features. 
To align $\mathbf{Q}_{\boldsymbol{\phi}, \mathbf{u}k}$ with $\mathbf{P}_{\boldsymbol{\phi}, \mathbf{u}_k}$, we minimize their KL divergence:
	\begin{equation}
		\text{KL}(\mathbf{P}_{\boldsymbol{\phi}, \mathbf{u}_k} \| \mathbf{Q}_{\boldsymbol{\phi}, \mathbf{u}_k};k) = \sum_{c=1}^{4C_k} \mathbf{P}_{\boldsymbol{\phi},\mathbf{u}_k}\left(c\right) \ln \frac{\mathbf{P}_{\boldsymbol{\phi},\mathbf{u}_k}\left(c\right)}{\mathbf{Q}_{\boldsymbol{\phi},\mathbf{u}_k}\left(c\right)}, \label{KLdivergence}
	\end{equation}
	where $\mathbf{P}_{\boldsymbol{\phi},\mathbf{u}_k}\left(c\right)$ and $\mathbf{Q}_{\boldsymbol{\phi},\mathbf{u}_k}\left(c\right)$ represent the $c$-th value of $\mathbf{P}_{\boldsymbol{\phi},\mathbf{u}_k}$ and $\mathbf{Q}_{\boldsymbol{\phi},\mathbf{u}_k}$, respectively.
	
In fact, this KL term prevents $\mathbf{Q}_{\boldsymbol{\phi},\mathbf{u}k}$ from becoming uniform over channels. 
By focusing the $\mathcal{G}$'s training on maximizing the dataset's likelihood, i.e., $-\log q_{\mathbf{\boldsymbol{\theta}}}(\mathbf{x})$, without the influence of the KL loss, each element in $\mathbf{P}_{\boldsymbol{\phi},\mathbf{u}_k}$ can reflect the varying importance of channel features. Consequently, $\mathbf{Q}{\boldsymbol{\phi},\mathbf{u}_k}$ also remains non-uniform across channels. 
	
A well-trained $\mathcal{G}$ is crucial for guiding $\mathcal{S}$, enabling effective shuffling and preserving feature information across scales. This refines the latent variables, improving their alignment with a normal distribution, thereby enhancing likelihood estimation. Section 4 further justifies this approach from an information-theoretic perspective.
	
In practice, KL divergence is estimated via Monte Carlo sampling of $\mathbf{u}_k$ by drawing $\mathbf{x}$ from $\mathcal{X}$. The total KL divergence is incorporated into the objective function:
\begin{equation}
    \begin{aligned}
        \mathcal{L}(\mathbf{x}; \boldsymbol{\theta},\boldsymbol{\phi}) &= \mathbb{E}_{\mathbf{x}\sim p(\mathbf{x})} \left[-\log q_{\mathbf{\boldsymbol{\theta}}}(\mathbf{x})  + \lambda \sum_{k=1}^{K-1} \text{KL}(\mathbf{P}_{\boldsymbol{\phi},\mathbf{u}_k} \| \mathbf{Q}_{\boldsymbol{\phi},\mathbf{u}_k};k) \right] \\
        &\approx \frac{1}{|\mathcal{X}|}\sum_{\mathbf{x} \in \mathcal{X}} \left[ -\log q_{\mathbf{\boldsymbol{\theta}}}(\mathbf{x})  + \lambda  \sum_{k=1}^{K-1} \sum_{c=1}^{4C_k} \mathbf{P}_{\boldsymbol{\phi},\mathbf{u}_k}\left(c\right) \ln \frac{\mathbf{P}_{\boldsymbol{\phi},\mathbf{u}_k}\left(c\right)}{\mathbf{Q}_{\boldsymbol{\phi},\mathbf{u}_k}\left(c\right)} \right], \\
        \label{eq:final-loss}
    \end{aligned}
\end{equation}
where $\boldsymbol{\theta}$ and $\boldsymbol{\phi}$ denote the parameters of the vanilla flow-based model and the parameters of $\mathcal{S}$ along with $\mathcal{G}$ introduced by our method, respectively. $\lambda$ balances the two losses. $|\mathcal{X}|$ denotes the set size of $\mathcal{X}$.
The proposed training process is shown in \cref{algorithm1}.
\begin{algorithm}[ht]
    \renewcommand{\algorithmicrequire}{\textbf{Input:}}
    \renewcommand{\algorithmicensure}{\textbf{Output:}}
    \caption{The training process for EIW-Flow}
    \label{algorithm1}
    \begin{algorithmic}
        \REQUIRE datapoint $\mathbf{x}_0$ from dataset $\mathcal{X}$, epoch number $J$, scale number $K$, steps of  flow $N$, hyperparameter $\lambda$.
        \STATE \textbf{Initialize:} the parameters $\boldsymbol{\theta}$ of vanilla model, the parameters $\boldsymbol{\phi}$ of \emph{solver}-$\mathcal{S}$ and \emph{guider}-$\mathcal{G}$ , total loss $\mathcal{L}=0$.
        \FOR {$j=1$ \TO $J$}
        \FOR {$k=1$ \TO $K$}
        \STATE \%\% \textbf{Step 1: Shuffle and Split latent variables}
        \STATE $\mathbf{u} _k = \mathtt{Squeeze}(\mathbf{f}_{k}^{N}\circ\cdots \circ\mathbf{f}_{k}^{2}\circ\mathbf{f}_{k}^{1}(\mathbf{x}_{k-1}))$
        \STATE $ \mathbf{Q}_{\boldsymbol{\phi},\mathbf{u} _k} \gets \mathcal{S}(\mathbf{u} _k, \boldsymbol{W}_{\mathcal{S}})$  $\;$ 
        \STATE $\hat{\mathbf{u}}_k\gets \mathcal{S}_{\mathcal{F}}(\mathbf{u}_k,\mathbf{Q}_{\boldsymbol{\phi},\mathbf{u} _k})$ 
        \STATE $[\mathbf{x}_k, \mathbf{z}_k] \gets \mathtt{Split}(\hat{\mathbf{u}}_k)$ 
        \STATE \%\% \textbf{Step 2: Calculate the KL divergence}
        \STATE $\mathbf{P}_{\boldsymbol{\phi},\mathbf{u} _k}\gets\mathcal{G}\left(\mathbf{u} _k, \mathbf{W}_{\mathcal{G}}\right)$ 
        \STATE $\mathcal{L}(\mathbf{x}_0;\boldsymbol{\theta},\boldsymbol{\phi}) \gets \mathcal{L}(\mathbf{x}_0;\boldsymbol{\theta},\boldsymbol{\phi}) + \lambda  \cdot \text{KL}(\mathbf{P}_{\boldsymbol{\phi},\mathbf{u} _k} \| \mathbf{Q}_{\boldsymbol{\phi},\mathbf{u} _k};k)$  $\;$ 
        \ENDFOR
        \STATE $\mathcal{L}(\mathbf{x}_0;\boldsymbol{\theta},\boldsymbol{\phi}) \gets \mathcal{L}(\mathbf{x}_0;\boldsymbol{\theta},\boldsymbol{\phi})-\log q_{\boldsymbol{\theta}}(\mathbf{x}_0)$   $\;$ %
        \STATE $[\boldsymbol{\theta}, \boldsymbol{\phi}] \gets \text{Adam}(\nabla_{\boldsymbol{\theta}}\mathcal{L}(\mathbf{x}_0;\boldsymbol{\theta},\boldsymbol{\phi}), \nabla_{ \boldsymbol{\phi}}\mathcal{L}(\mathbf{x}_0;\boldsymbol{\theta},\boldsymbol{\phi}), \boldsymbol{\theta},\boldsymbol{\phi})$  $\;$ 
        \ENDFOR
        \ENSURE the trained parameters $\boldsymbol{\theta}$ and $\boldsymbol{\phi}$.
    \end{algorithmic}
\end{algorithm}

\section{Theorem Analysis}
\label{sec:analysis}
	
In \cref{sec:method}, we introduced the $\mathtt{Shuffle}$ operation, and its effectiveness will be empirically demonstrated in \cref{sec:quantitative-evaluation} and \cref{sec:qualitative-evaluation}. However, the theoretical reasons behind its efficacy remain unclear. In this section, we analyze $\mathtt{Shuffle}$ from an information-theoretic perspective.
We demonstrate that $\mathtt{Shuffle}$ increases the entropy difference between $\{\mathbf{x}_k\}_{k=1}^{K-1}$ and $\{\mathbf{z}_k\}_{k=1}^{K-1}$ under the EIW-Flow framework compared to the vanilla model. First, we qualitatively analyze its entropy-increasing mechanism using \cref{theorem:expected-entropy-increase} and \cref{proposition:R2E2-compare}. Then, we quantitatively validate these findings through ablation studies on CIFAR-10, CelebA, and MNIST.

\subsection{Theoretical Foundations}
\textbf{Information bottleneck principle.}
The $\mathtt{Shuffle}$ aligns with the information bottleneck framework \cite{zhang1997non}, which aims to compress input data while retaining essential information. Here, $\mathbf{z}_k$ acts as a bottleneck, discarding redundant features, while $\mathbf{x}_k$ preserves high-fidelity information. By optimizing $\mathbf{Q}_{\boldsymbol{\phi},\mathbf{u}k}$, $\mathtt{Shuffle}$ ensures $\mathbf{z}_k$ primarily contains redundant features, leading to smaller information loss (see \cref{theorem:information-loss-reduction}).

\textbf{Manifold topology preservation.}
High-dimensional data often lie on low-dimensional manifolds. 
Static splitting used in previous works disrupts this structure by arbitrarily removing channels.
In contrast, $\mathtt{Shuffle}$ preserves the manifold’s topology by retaining channels critical to its intrinsic geometry (guaranteed by \emph{guider}-$\mathcal{G}$).
This aligns with manifold learning theory \cite{roweis2000nonlinear}, where adaptive feature selection maintains the data’s intrinsic structure.

\textbf{Central Limit Theorem and Maximum Entropy Principle.}
NFs map complex data distributions to a standard Gaussian, a process supported by the Central Limit Theorem (CLT) \cite{rosenblatt1956central} and the Maximum Entropy Principle (MEP) \cite{shore1980axiomatic}. These principles imply that well-trained NFs increase entropy and uncertainty.
	
\begin{theorem}[Maximum Entropy Principle, MEP\cite{shore1980axiomatic}]
\label{theorem:maximum-entropy-principle}
Let $\mathbf{x}$ be a random variable with probability density $p(\mathbf{x})$. Under constraints on its mean and variance, the distribution maximizing entropy $H(p)$ is the standard Gaussian. 
\end{theorem}

\subsection{Adaptive Feature Propagation}
Traditional splitting methods divide latent features into fixed groups, ignoring channel importance and leading to irreversible information loss. For instance, in image generation, critical features like facial contours might be randomly discarded (see \cref{fig:celeba-middle-status-train-vanilla} for visualization).

Our proposed $\mathtt{Shuffle}$, an adaptive feature propagation mechanism, can mitigate this problem with smaller information loss. Let $\mathbf{Q}_{\boldsymbol{\phi},\mathbf{u}_k} \in \mathbb{R}^{4C_k}$ denote learned channel importance scores. Channels are ranked by $\mathbf{Q}_{\boldsymbol{\phi},\mathbf{u}_k} \in \mathbb{R}^{4C_k}$, with the top $2C_k$ allocated to $\mathbf{x}_k$, while the rest to $\mathbf{z}_k$.

\begin{theorem}[Information Loss Reduction]
\label{theorem:information-loss-reduction}
Given a latent variable $\mathbf{u}_k$,
let $\mathcal{L}_{\mathrm{static}} = I(\mathbf{u}_k; \mathbf{x}_k) - I(\mathbf{u}_k; \mathbf{z}_k)$ and $\mathcal{L}_{\mathrm{adaptive}}= I(\mathbf{u}_k; \mathbf{x}_k | \mathbf{Q}_{\boldsymbol{\phi},\mathbf{u}_k}) -  I(\mathbf{u}_k; \mathbf{z}_k | \mathbf{Q}_{\boldsymbol{\phi},\mathbf{u}_k})$ be the information losses without and with $\mathtt{Shuffle}$ operation, where $I(\mathbf{u}_k; \mathbf{x}_k)$ measures how much useful information is retained, and $I(\mathbf{u}_k; \mathbf{z}_k)$ quantifies information loss in discarded channels. Then, the inequality $
\mathcal{L}_{\mathrm{adaptive}} \leq \mathcal{L}_{\mathrm{static}}$ holds for all $k$.
\end{theorem}

\noindent \textit{Proof.} The $\mathtt{Shuffle}$ operation guarantees reduced information loss compared to static splitting used in previous works, i.e., $
\mathcal{L}_{\text{adaptive}} \leq \mathcal{L}_{\text{static}}$. To prove this, we first construct a Markov chain $\mathbf{u}_k \rightarrow \mathbf{Q}_{\boldsymbol{\phi},\mathbf{u}_k} \rightarrow \hat{\mathbf{u}}_k$, where  $\hat{\mathbf{u}}_k$ is the shuffled feature tensor. By the Data Processing Inequality \cite{zhang1997non}, we have $I(\mathbf{u}_k; \hat{\mathbf{u}}_k) \leq I(\mathbf{u}_k; \mathbf{Q}_{\boldsymbol{\phi},\mathbf{u}_k})$, meaning the shuffled features retain no more information than the scores themselves. Critically, by sorting channels based on $\mathbf{Q}_{\boldsymbol{\phi},\mathbf{u}_k}$, $\mathtt{Shuffle}$ ensures $\mathbf{x}_k$ maximizes $I(\mathbf{u}_k; \mathbf{x}_k)$ (retaining high-importance channels) while minimizing $I(\mathbf{u}_k; \mathbf{z}_k)$. Substituting these into $\mathcal{L}_{\text{static}}$, we directly derive $\mathcal{L}_{\text{adaptive}} \leq \mathcal{L}_{\text{static}}$.  $\Box$

\subsection{Entropy Increase Process}	
To analyze the nature of our $\mathtt{Shuffle}$ operation from the  perspective of entropy theory, we design an ablation experiment on CIFAR-10, CelebA and MNIST datasets.
We firstly approximate the entropy of each element of $\mathbf{x}_k$ and $\mathbf{z}_k$ 
$(k=1,2,\ldots, K-1)$ with Monte Carlo Estimation, namely:
\begin{equation}
        H(p) = -\int_{\mathbb{R}}p(\mathbf{x})\log p(\mathbf{x})d \mathbf{x} \approx -\frac{1}{|\mathcal{X}|}\sum_{\mathbf{x} \in \mathcal{X}} \log p(\mathbf{x}).
\end{equation}
	
Then we calculate the mathematical expectation of the entropy for each element of $\mathbf{x}_k$ and $\mathbf{z}_k$, which we term as the expected entropy of $\mathbf{x}_k$ and $\mathbf{z}_k$. 

\begin{definition}[Expected Entropy, E2]
    Consider a random variable $\mathbf{x}$, its expected entropy is defined as follows:
    \begin{equation}
        \mathrm{E2}(\mathbf{x} ) \triangleq \frac{2}{C\times H \times W}\sum_{c=1}^{C/2}\sum_{a=1}^{H}\sum_{b=1}^{W}H(p(x^{c}(a,b))), \label{eq:entropy_vector}
    \end{equation}
    where $C$ and $[H,W]$ are the channel and spatial size of $\mathbf{x}$, $p(x^{c}(a,b))$ is the probability density distribution of $x^{c}(a,b)$, $x^{c}(a,b)$ denotes the value at $(a,b)$ of the $c$-th channel feature map of $\mathbf{x}$. A greater value of $\mathrm{E2}$ indicates that $\mathbf{x}$ is more likely to follow a standard Gaussian distribution.
\end{definition}

Furthermore, to better compare vanilla architecture with our model, we define a new indicator named Relative Ratio of Expected Entropy ($\mathrm{R2E2}$).

\begin{definition}[Relative Ratio of Expected Entropy, $\mathrm{R2E2}$]
    Consider two random variables $\mathbf{x}$ and $\mathbf{z}$, their Relative Ratio of Expected Entropy is defined as follows:
    \begin{equation}
        \mathrm{R2E2}(\mathbf{x}, \mathbf{z}) \triangleq \frac{\mathrm{E2}(\mathbf{z}) - \mathrm{E2}(\mathbf{x})}{\mathrm{E2}(\mathbf{x})}.
    \end{equation}
    where $\mathrm{E2}(\cdot)$ is the expected entropy function as defined in  \cref{eq:entropy_vector}. A greater value of $\mathrm{R2E2}$ indicates that the expected entropy of $\mathbf{z}$ is more greater than that of $\mathbf{x}$ without considering the magnitude, which also means $\mathbf{z}$ has less information.
\end{definition}
	
According to MEP, if $\mathbf{z}_k$ truly follows a standard Gaussian distribution as assumed in NFs, $\mathrm{E2}(\mathbf{z}_k)$ ought to be relatively greater, while the opposite is true for $\mathbf{x}_k$. In this case, the value of $\mathrm{R2E2}(\mathbf{x}_k, \mathbf{z}_k)$ should also be greater. 

\begin{theorem}
\label{theorem:expected-entropy-increase}
    The expected entropy of latent variable increases during the inference process of EIW-Flow.   
\end{theorem}
	
\noindent \textit{Proof.} EIW-Flow serves as a feature information extractor,  progressively extracting feature information from input data. This results in the final latent variable follows a standard Gaussian distribution. Moreover, this also leads to an increase in the entropy value of latent variables in accordance with MEP. Specifically, the entropy of each element of the latent variables, such as $H(p(x^{c}(a,b)))$, also grows with the increase in scale number until they follow standard Gaussian distribution. Consequently, each term of the right hand side of \cref{eq:entropy_vector} increases during the inference process. As a result, the expected entropy of latent variables naturally increases throughout the inference process according to the definition of expected entropy. $\Box$
	
\begin{proposition}
\label{proposition:R2E2-compare}
    For $\forall k \in \{1,2,\ldots,K-1 \}$, the $\mathrm{R2E2}$ between $\mathbf{x}_{k}$ and $\mathbf{z}_{k}$ for EIW-Flow is greater than that for vanilla normalizing flows.
\end{proposition}
	
\noindent \textit{Proof.} As described in \cref{solver}, the $\mathtt{Shuffle}$ operation propagates relatively information-rich channel feature maps of latent variables like $\mathbf{x}_{k}$ to the next scale. These channel feature maps have smaller entropy value based on MEP, indicating that $H(p(\mathbf{x}_{k}))$ is smaller than $H(p(\mathbf{z}_{k}))$ for all $k \in \{1,2,\ldots,K-1 \}$. Additionally, some channel feature maps of $\mathbf{z}_{k}$ may contain more feature information of input data than that of $\mathbf{x}_{k}$ as shown in \cref{fig:celeba-middle-status-train-vanilla} when the $\mathtt{Shuffle}$ operation is not applied. This implies the difference between $\mathrm{E2}(\mathbf{z}_{k})$ and $\mathrm{E2}(\mathbf{x}_{k})$ is greater for EIW-Flow compared to the vanilla model, resulting in a greater $\mathrm{R2E2}(\mathbf{x}_{k}, \mathbf{z}_{k})$. $\Box$
	
\cref{proposition:R2E2-compare} states that channels with higher expected entropy are selected to form the final latent variable in EIW-Flow, i.e., $\mathrm{E2}(\mathbf{z}_{k}) > \mathrm{E2}(\mathbf{x}_{k})$ for all $k$. 
According to \cref{theorem:maximum-entropy-principle}, a variable containing richer and more important features tends to have lower entropy \cite{zhang2016feature}. 
The elements of $\mathbf{P}_{\boldsymbol{\phi}}$  are positively correlated with feature richness and importance, meaning higher values correspond to lower entropy \cite{chen2025dequantified}. 
By optimizing with \cref{KLdivergence},  $\mathbf{Q}_{\boldsymbol{\phi}}$  from \emph{solver}-$\mathcal{S}$ progressively aligns with $\mathbf{P}_{\boldsymbol{\phi}}$, naturally propagating higher-entropy elements to form the final latent variable. Experimental results in \cref{tab:entropy-increment} validate this proposition.
\begin{table}[ht]
    \caption{Expected Shannon information entropy before (vanilla) and after (our) our $\mathtt{Shuffle}$ operation on three image datasets. }
    
    \centering
    \resizebox{\textwidth}{!}{
        \begin{tabular}{llccc|ccc|c}
            \toprule
            dataset& stage &vanilla $\mathbf{x}_k$ & vanilla $\mathbf{z}_k$ & \text{R2E2}\textuparrow & our $\mathbf{x}_k$ & our $\mathbf{z}_k$ & \text{R2E2}\textuparrow & improvement rate\\
            \midrule
            CIFAR-10 ($K=3$) &$k=1$ (training) &126.686 & 95.716 & -0.244 & 91.616 & 253.140 & \textbf{1.763} &82.25\% \textuparrow\\
            &$k=2$ (training) &160.136 & 198.423 & 0.239 & 145.897 & 257.104 & \textbf{0.762}& 21.88\% \textuparrow\\
            &$k=1$ (test) &112.938 & 122.165 & 0.082 & 103.350 & 205.970 & \textbf{0.993} & 111.1\% \textuparrow\\
            &$k=2$ (test) &154.229 & 183.434 & 0.189 & 139.671 & 209.362 & \textbf{0.499} & 16.40\% \textuparrow\\
            \hline
            CelebA ($K=3$)  &$k=1$ (training) &82.785 & 68.522 & -0.172 & 62.996 & 184.765 & \textbf{1.933} & 122.4\% \textuparrow\\
            &$k=2$ (training) &140.938 & 188.675 & 0.339 & 123.909 & 219.719 & \textbf{0.773} & 12.80\% \textuparrow\\
            &$k=1$ (test) &80.844 & 63.133 & -0.219 & 59.192 & 193.134 & \textbf{2.263} & 113.3\% \textuparrow\\
            &$k=2$ (test) &144.818 & 175.831 & 0.214 & 127.732 & 208.120 & \textbf{0.629} & 19.39\% \textuparrow\\
            \hline
            MNIST ($K=2$)   &$k=1$ (training) &159.195 & 163.357 & 0.026 & 155.329 & 162.525 & \textbf{0.046} & 7.692\% \textuparrow\\
            &$k=1$ (test) &135.019 & 144.364 & 0.069 & 129.791 & 166.534 & \textbf{0.283} & 31.01\% \textuparrow\\
            \bottomrule
        \end{tabular}
    }
    \label{tab:entropy-increment}
\end{table}
	
From \cref{tab:entropy-increment}, we observe: 1) Across all datasets, the \text{R2E2} significantly increases after applying $\mathtt{Shuffle}$; 
2) in our model, $\mathrm{E2}(\mathbf{z}_{k}) > \mathrm{E2}(\mathbf{x}_{k})$, while the vanilla model has some outliers; 
3) both $\mathrm{E2}(\mathbf{z}_{k})$ and $\mathrm{E2}(\mathbf{x}_{k})$ increase with $k$. Combining (1) and (2), our $\mathtt{Shuffle}$ operation effectively propagates high-entropy channel feature maps, which better approximate a Gaussian distribution, to the final latent variable. As for (3), under NF assumptions, $\mathbf{z}_k$ follows a normal distribution while $\mathbf{x}_k$ follows a complex, unknown distribution. 
Our experiments demonstrate that the shuffle operation provides more significant improvements on complex datasets (e.g., CelebA) than on simpler ones (e.g., MNIST), as it more effectively accelerates the transformation of intricate data distributions to normal distributions within the same network depth.
These results align with NF theoretical expectations.
\begin{figure}[ht]
    \centering
    \includegraphics[width=0.8\linewidth]{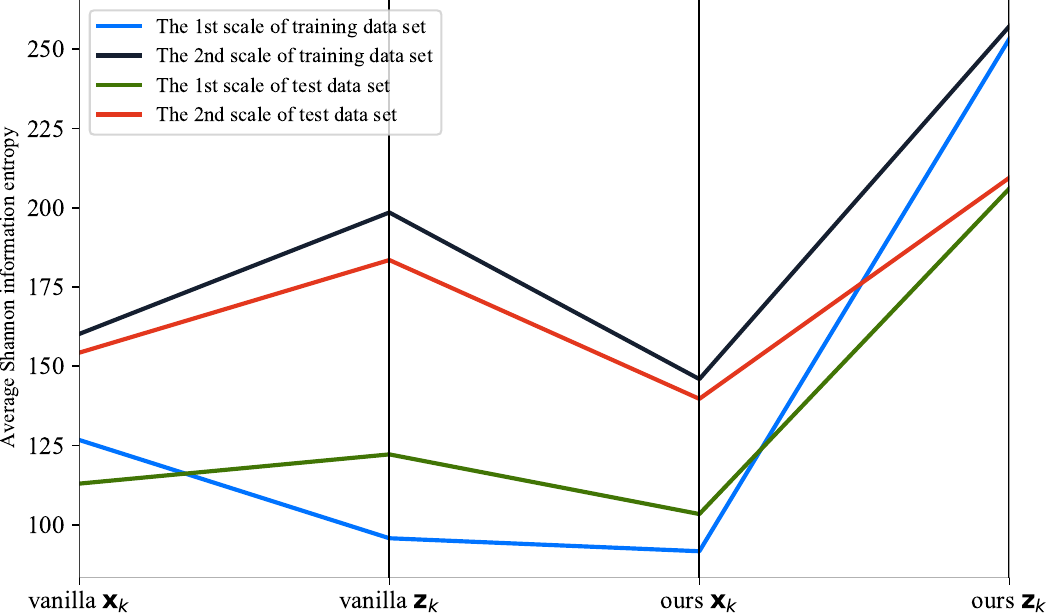}
    \caption{Comparative analysis of average Shannon information entropy across scales and methods.}
    \label{fig:parallelcompareentropy}
\end{figure}
	
Our quantitative evaluation reveals three key findings about the entropy distributions: First, Kolmogorov-Smirnov tests confirm our method successfully produces more Gaussian-distributed $\mathbf{z}$-features (average KS$=0.12\pm0.03$ across scales) compared to vanilla's $\mathbf{z}$-features (KS$=0.91\pm0.05$), while deliberately enhancing non-Gaussian characteristics in $\mathbf{x}$-features (ours KS$=0.85\pm0.04$ vs vanilla KS$=0.79\pm0.06$). Second, the parallel coordinates plot (see \cref{fig:parallelcompareentropy}) demonstrates our method's consistent superiority in $\mathbf{z}$-dimension entropy across all scales, with all improvements statistically significant ($p<0.001$, bootstrap-tests). Third, the controlled $\mathbf{x}$-dimension entropy reduction maintains clearer separation from $\mathbf{z}$-features than vanilla's approach.
	
\section{Quantitative Evaluation}
\label{sec:quantitative-evaluation}

\subsection{Experimental Setup}
We implemented dataset-specific configurations for all experiments: For CelebA, Imagenet64, LSUN Church, and CelebA-HQ, we trained with a batch size of $16$ for $70$ epochs using the Adamax optimizer (learning rate=$1e-4$, $\beta_1=0.9$, $\beta_2=0.999$). For CIFAR-10, MNIST and Imagenet32, we increased the batch size to 64 and extended training to $650$ epochs with a lower learning rate with $2e-5$ while maintaining the same optimizer settings, and applied a learning rate schedule with $0.9975$. All models were trained with 8-bit quantization and dataset-specific augmentations - horizontal flipping ($p=0.5$) for CelebA-family datasets. The complete hyperparameter configurations are provided in our released code repository:  \url{https://github.com/studying910/EIW-Flow/tree/main/DIW-Flow}.
	
\subsection{Density Estimation}
In this section, we evaluate our model on three benchmark datasets: CIFAR-10, CelebA and ImageNet  (resized to $32\times32$ and $64\times64$ pixels). 
Since natural images are discretized, we apply a variational dequantization method \cite{ho2019flow++} to obtain continuous data necessary for NFs. We set the hyperparameter $\lambda=0.001$ to balance the optimization of the neural network $q_{\boldsymbol{\theta}}$ with the \emph{solver}-$\mathcal{S}$. A too large $\lambda$ will result in neglecting the optimization of backbone neural network $q_{\boldsymbol{\theta}}$, while a too small $\lambda$ will prevent the \emph{solver}-$\mathcal{S}$ from adequately approximating the \emph{guider}-$\mathcal{G}$ we introduced, thus degrading the model to the original architecture as described in  \cref{multiscale}. We set the number of scales $K=3$ and the number of transformations $N=6$ in each scale so that the model is expressive enough and does not take too long to train. We choose to use the Adam optimizer to make a fair comparison with other models. We take advantage of the warm-up procedure described in \cite{grcic2021densely}. All our experiments are conducted on a single Tesla-V100 GPU with $32$ GB memory.
	
\cref{sota} compares the generative performance of different NFs models. On these six datasets, our model achieves the best performance among NFs, which are $2.97$, $1.96$, $3.62$, $3.35$, 0.97, 2.31, respectively. The reported results are averaged over five runs with different random seeds.
\begin{table*}[ht]
    \caption[Negative log-likelihood (bits/dim)]{Negative log-likelihood (bits/dim) evaluation on image datasets. Asterisks (*) indicate that the model is best  among all Normalizing Flows models. A short horizontal line (-) indicates that the corresponding paper did not experiment with this dataset.}
    \centering
    \resizebox{\linewidth}{!}{
    \begin{tabular}{lcccccccc}
        \toprule
        \textbf{Model} & \textbf{CIFAR-10} & \textbf{CelebA} & \textbf{ImageNet32} & \textbf{ImageNet64} & \textbf{CelebA HQ} & \textbf{LSUN (church)}\\
        \midrule
        Residual Flow \cite{chen2019residual} & 3.28 & - & 4.01 & 3.78 & 0.992 & - \\
        DenseFlow \cite{grcic2021densely} & 2.98 & 1.99 & 3.63 & 3.35 & - & - \\
        Flow++ \cite{ho2019flow++} & 3.08 & - & 3.86 & 3.69 & - & - \\
        REAL NVP \cite{dinh2016density} & 3.49 & 3.02 & 4.28 & 3.98 & - & 3.08 \\
    GLOW \cite{kingma2018glow} & 3.35 & - & 4.09 & 3.81 & 1.03 & 2.67  \\
        Wavelet Flow \cite{yu2020wavelet} & - & - & 4.08 & 3.78 & - & 2.74\\
        EIW-Flow (ours) & \textbf{2.97}$^{*}$ & \textbf{1.96}$^{*}$ & \textbf{3.62}$^{*}$ & \textbf{3.35}$^{*}$ & \textbf{0.97}$^{*}$ & \textbf{2.31}$^{*}$\\
        \hline
    \end{tabular}
    }
    \label{sota}
    \centering
\end{table*}
	
\subsection{Experiments on MNIST dataset}
Additionally, we conduct quantitative and qualitative experiments with the MNIST dataset, which is a simple yet meaningful dataset. Different from other datasets, we set the number of scale $K=2$ and the number of transformations $N=6$. The other hyperparameters and settings are the same as other three datasets.

We compare EIW-Flow with other NFs according to bits/dim and FID metrics. Results are presented in  \cref{bpd-and-fid}. Our model achieves the best performance among NFs for both metrics, which are $0.89$ and $2.4$, respectively. This shows that our model is more suitable for generating high-fidelity handwritten digital images than other NFs.
\begin{table}[ht]
    \caption{Evaluation of bits/dim and FID score on MNIST dataset. The "$\star$" indicates that the model is optimal for all models compared. The "-" indicates that this model was not experimented on this dataset in the corresponding paper.}
    \centering
    \resizebox{\linewidth}{!}{
    \begin{tabular}{lcc|lcc}
        \toprule
        \textbf{Model} & \textbf{bits/dim}$\downarrow $ & \textbf{FID}$\downarrow $ & \textbf{Model} & \textbf{bits/dim}$\downarrow $ & \textbf{FID}$\downarrow $  \\
        \midrule
        Residual Flow \cite{chen2019residual} & 0.97 & - &  SINF \cite{dai2020sliced} & 1.34 & 4.5 \\
        TO-Flow \cite{du2022flow} & 1.03 & - & GLF+perceptual loss \cite{xiao2019generative} & - & 5.8\\
        i-ResNet \cite{behrmann2019invertible} & 1.06 & - & EIW-Flow (ours) & \textbf{0.89}$^{\star}$ & \textbf{2.4}$^{\star}$ \\
        \bottomrule
    \end{tabular}
    }
    \label{bpd-and-fid}
\end{table}
	
We also generate handwritten digital images from our trained model by sampling $\mathbf{z} \sim N(\mathbf{0}, \mathbf{I})$ in the latent space and reversing $\mathbf{z}$ according to the sampling process of EIW-Flow. The sampling process is described in \cref{visualquality} and the results are shown in \cref{fig:mnist-sample}. We can observe that the images in  \cref{fig:mnist-sample} have clear structures.

\subsection{Computational Complexity}
Since we have introduced a \emph{solver}-$\mathcal{S}$ to automatically assign weights to channels at each scale, additional computational overhead will be inevitably added and thus limit the inference speed. Therefore, we use frames per second (FPS) on all four datasets to compare our trained model with vanilla model using the original architecture. For CIFAR-10 and ImageNet32, we set the batch size to $64$, while in other two cases it is set to $16$ due to memory limitations. The detailed results are shown in \cref{tab:time-memory}. The reported results are obtained on a single Tesla V-100 GPU. 
\begin{table}[htbp]
\centering
\caption{Comparison of Frames Per Second (FPS), parameter size, training time and MACs. The reported results are obtained on a single Tesla V-100 GPU with batch size $16$.}
\setlength{\tabcolsep}{2.3pt}
\resizebox{\linewidth}{!}{
\begin{tabular}{c|cc|cc|cc|cc|cc}
    \hline
    \multirow{3}{*}{\textbf{Dataset}}  & \multicolumn{4}{c|}{\textbf{FPS}$\uparrow$} & \multicolumn{2}{c|}{\textbf{Param Size (M)}$\downarrow$} & \multicolumn{2}{c|}{\textbf{Duration (h)}$\downarrow$} & \multicolumn{2}{c}{\textbf{MACs (G)}$\downarrow$} \\
    \cline{2-11}          & \multicolumn{2}{c|}{Training set} & \multicolumn{2}{c|}{Test set} & \multirow{2}{*}{our} & \multirow{2}{*}{vanilla} & \multirow{2}{*}{our} & \multirow{2}{*}{vanilla} & \multirow{2}{*}{our} & \multirow{2}{*}{vanilla} \\
    \cline{2-5}     & our   & vanilla & our   & vanilla & & & & & &  \\
    \hline 
    \textbf{CIFAR-10} & $1103$  & $1088$  & $1103$  & $1088$  & $130.92$ & $130.65$ & $390$ & $400$ & $14.98$ & $14.98$ \\
    \textbf{CelebA} & $24$    & $25$    & $22$    & $23$  & $130.92$ & $130.65$ & $375$ & $360$ & $59.93$ & $59.93$ \\
    \textbf{ImageNet32} & $102$ & $105$ & $88$    & $90$ & $130.92$ & $130.65$ & $500$ & $490$ & $14.98$ & $14.98$ \\
    \textbf{ImageNet64} & $24$    & $25$    & $24$    & $25$    & $130.92$ & $130.65$ & $2010$ & $1990$ & $59.93$ & $59.93$ \\
    \hline
\end{tabular}%
}
\label{tab:time-memory}%
\end{table}%

For CelebA and ImageNet64 datasets, our improved model adds less than $0.03$ seconds to vanilla one, which is insignificant in relation to the total elapsed time. As for the other two cases, we can observe that the increase in total elapsed time for EIW-Flow is negligible compared to the vanilla model. Moreover, the  number of parameters of our model and vanilla model are $130.92$ million and $130.65$ million, respectively. Only $270$k parameters are added by \emph{solver}-$\mathcal{S}$, indicating that the improvement in expressive power does not come from the deepening of our model.

\section{Qualitative Evaluation}
\label{sec:qualitative-evaluation}
\subsection{Visual Quality}
\label{visualquality}
The analysis of visual quality is important as it is well-known that calculating log-likelihood is not necessarily indicative of visual fidelity. In this part, we use the FID metric \cite{heusel2017gans} to evaluate the visual quality of the samples generated by EIW-Flow. The FID score requires a large corpus of generated samples in order to provide an unbiased estimation. Hence, due to the memory limitation of our platform, we generate $50$k samples for all two datasets, which is also the default number in the calculation of FID score. Corresponding results can be seen in  \cref{fid}.
\begin{table*}[ht]
    \caption{Evaluation of FID score on two image datasets. Lower is better. Asterisks ($\star$) indicate that the model is best  among all Normalizing Flows models. A short horizontal line (-) indicates that the corresponding paper did not experiment with this dataset. We did not test it on ImageNet64 since the number of images needed to be stored exceed the memory limit.}
    
    \centering
    \begin{tabular}{lcccccc}
        \toprule
        \textbf{Model} & \textbf{CIFAR-10} & \textbf{CelebA} & \textbf{ImageNet32} 
        \\
        \midrule
        GLOW \cite{kingma2018glow} & 46.90 & 23.32 & - \\
        DenseFlow \cite{grcic2021densely} & 34.90 & \textbf{17.1}$^{\star}$ & 38.5\\
        FCE \cite{gao2020flow} & 37.30 & 12.21 & - \\
        EIW-Flow (ours) & \textbf{31.60}$^{\star}$ & 18.0 & \textbf{36.5}$^{\star}$ \\
        \bottomrule
    \end{tabular}
    \label{fid}
\end{table*}

The generated ImageNet32 samples achieve a FID score of $36.5$, the CelebA samples achieve $18.0$ and  when using the training dataset. When using the validation dataset, we achieve $34.3$ on CIFAR-10, $18.0$ on CelebA and $37.5$ on ImageNet32. Our model outperforms the majority of Normalizing Flows.

\begin{figure}[!ht]
    \centering
        \subfloat[LSUN Church ($256\times256$ pixels).]{\label{fig:church-sample}\includegraphics[width=0.85\linewidth]{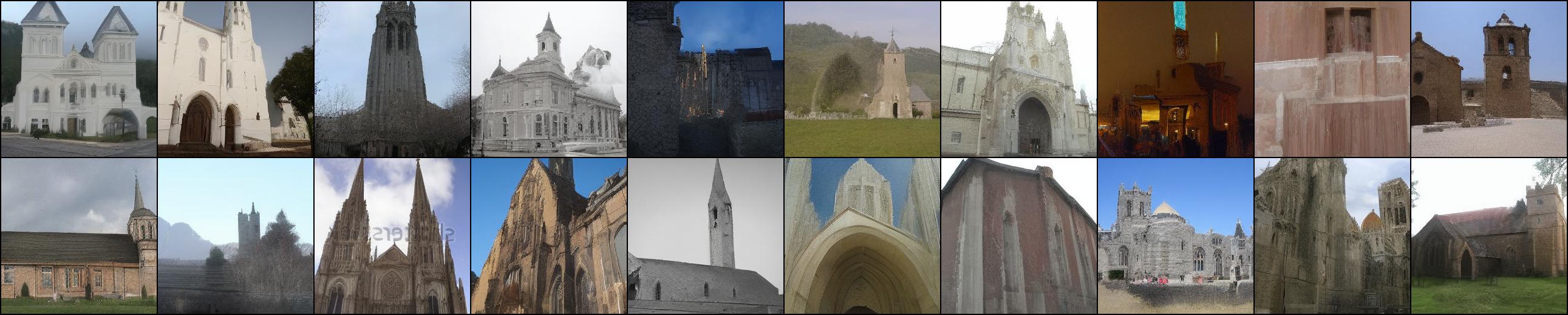}} \\
        \subfloat[CelebA-HQ ($256\times256$ pixels).]{\label{fig:celeba-hq-sample}\includegraphics[width=0.85\linewidth]{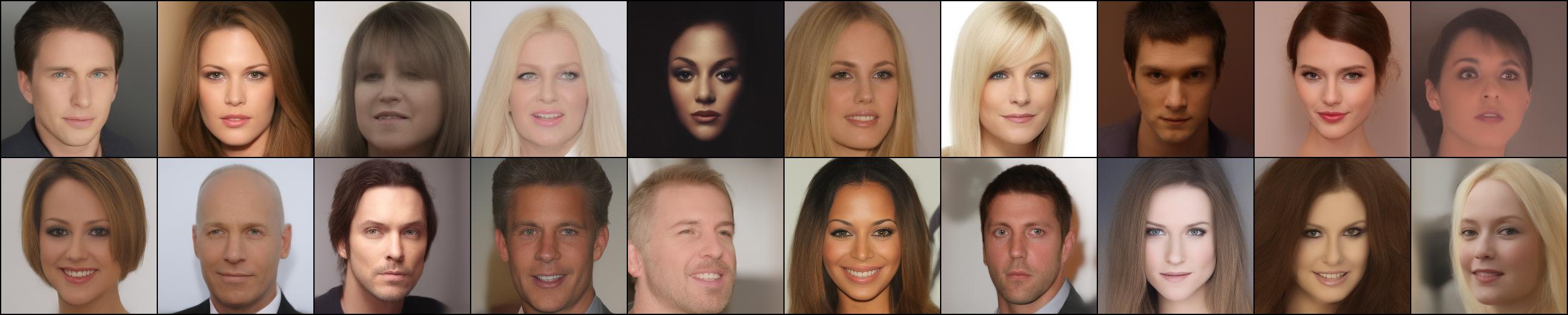}} \\
    \subfloat[CelebA ($64\times64$ pixels).]{\label{fig:celeba-sample}\includegraphics[width=0.85\linewidth]{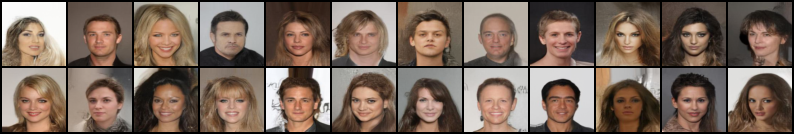}} \\
    \subfloat[ImageNet ($64\times64$ pixels).]{\label{fig:imn64-sample}\includegraphics[width=0.85\linewidth]{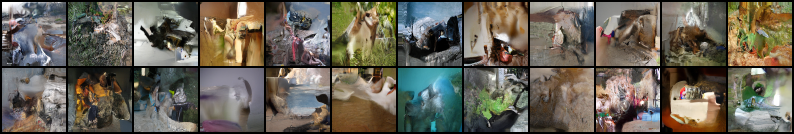}}\\
    \subfloat[CIFAR-10 ($32\times32$ pixels).]{\label{fig:cifar10-sample}\includegraphics[width=0.85\linewidth]{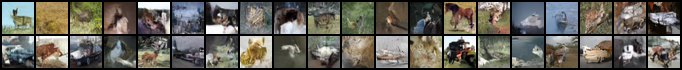}}
    \\
    \subfloat[MNIST ($28\times28$ pixels).]{\label{fig:mnist-sample}\includegraphics[width=0.85\linewidth]{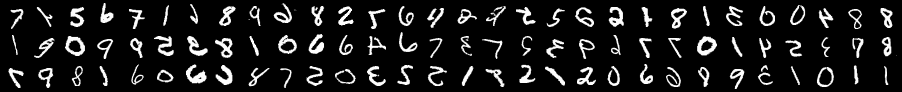}}
    \caption{Random samples generated by EIW-Flow on six image datasets. The temperature factor of Church, CelebA-HQ, CelebA, CIFAR-10 and MNIST are set to $0.8, 0.8, 0.8, 0.95$ and $0.95$, respectively.}
    \label{fig:sample}
\end{figure}

We also samples from the trained model through a four-step iterative process. At the $k$-th scale, we (1) sample $\mathbf{z}_{k}\sim \mathcal{N}(\boldsymbol{0}, \mathbf{I})$ with the same shape as  $\mathbf{x}_{k}$; (2) concatenate $\mathbf{z}_{k}$ with $\mathbf{x}_k$ to form $\hat{\mathbf{u}}_{k}$; (3) apply the inverse $\mathtt{Shuffle}$ operation followed by an $\mathtt{Unsqueeze}$ operation, which is the inversion of the $\mathtt{Squeeze}$ operation; and (4) pass the result $\mathbf{u}_{k}$ through the inverse flow steps $\{\mathbf{f}_k^n\}_{n=1}^{N}$ to obtain $\mathbf{x}_{k-1}$.
The complete sampling process at the $k$-th scale is as follows:
\begin{align}
    \mathbf{z}_k &\sim \mathcal{N}(\mathbf{0},\mathbf{I}), \label{sample} \\
    \hat{\mathbf{u}}_k &= \mathtt{Concat}(\mathbf{x}_k, \mathbf{z}_k), \label{concat-sample} \\
    \mathbf{u}_{k} &= \mathtt{Shuffle}^{-1}(\hat{\mathbf{u}}_{k}), \label{inverse-rem} \\
    \mathbf{x}_{k-1} &= \mathbf{g}_{k}^{1}\circ\mathbf{g}_{k}^{2}\circ\cdots \circ\mathbf{g}_{k}^{N}(\mathtt{Unsqueeze}(\mathbf{u}_{k})), \label{unsqueeze} 
\end{align}
where $\mathtt{Concat}$ and $\mathtt{Unsqueeze}$ are the inverses of $\mathtt{Split}$ and $\mathtt{Squeeze}$ (\cref{multiscale}). Similarly, $\mathtt{Shuffle}^{-1}$ and $\mathbf{g}_{k}^{n}\triangleq(\mathbf{f}_{k}^{n})^{-1}$ denote the inverses of $\mathtt{Shuffle}$ and  $\mathbf{f}_{k}^{n}$ (\cref{fig:diagramofselect}, \cref{multiscale}).  Sample results on LSUN Church, CelebA-HQ, CelebA, ImageNet, CIFAR-10 and MNIST are shown in \cref{fig:sample}. 

\begin{figure}[!ht]
    \centering
    \subfloat[DDPM-IP \cite{ning2023input}.]{\label{fig:celeba-DDPM-sample}\includegraphics[width=0.45\linewidth]{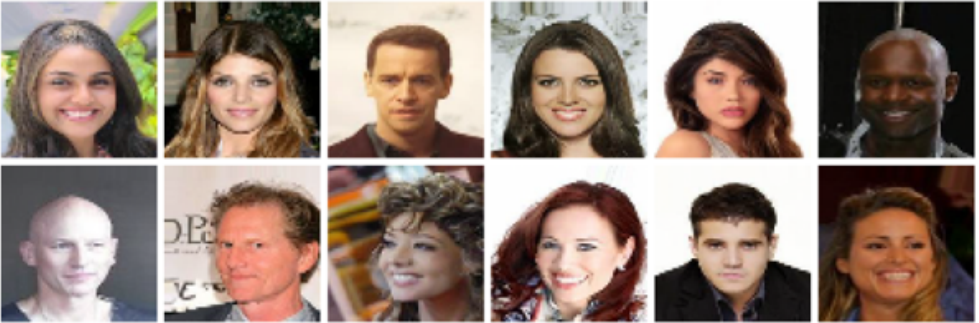}} 
    \subfloat[EIW-Flow (ours).]{\label{fig:celeba-ours-sample}\includegraphics[width=0.45\linewidth]{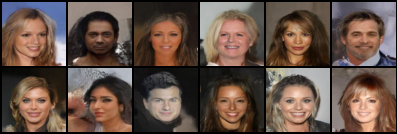}} 
    \caption{Comparisons between recent denoising diffusion probabilistic model and EIW-Flow (ours) on the CelebA dataset.}
    \label{fig:celeb-compare}
\end{figure}

We also compare EIW-Flow with recent denoising diffusion probabilistic model on CelebA, as presented in \cref{fig:celeb-compare}. For CelebA, we apply the reduced-temperature trick \cite{kingma2018glow} with the temperature factor $T=0.8$ to improve sample quality.

\subsection{Reconstruction}
\begin{figure}[!ht]%
    \centering
    \subfloat[CelebA ($64\times64$ pixels).]{\label{fig:celeba-test-reconstruction}\includegraphics[width=0.99\linewidth]{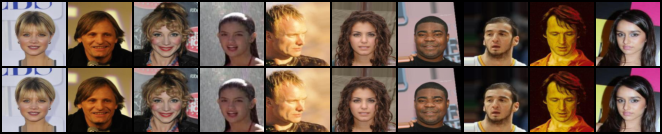}} \\
    \subfloat[CIFAR-10 ($32\times32$ pixels).]{\label{fig:cifar10-reconstruction}\includegraphics[width=0.99\linewidth]{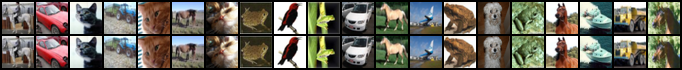}}%
    \caption{Reconstruction on CelebA and CIFAR-10. The first and second rows represent the original and the reconstructed images, respectively.}
    \label{fig:celeba-reconstruction}
\end{figure}

We first sample several datapoints $\mathbf{x}$ from the corresponding dataset and reconstruct them using $\hat{\mathbf{x}}=\mathbf{f}^{-1}(\mathbf{f}(\mathbf{x}))$, where $\mathbf{f}$ represents the forward transformation of our model. We sample from the training and test sets of CelebA and CIFAR-10, with results shown in \cref{fig:celeba-test-reconstruction} and \cref{fig:cifar10-reconstruction}. The top and bottom rows correspond to $\mathbf{x}$ and $\hat{\mathbf{x}}$, respectively. The near-identical reconstruction of $\hat{\mathbf{x}}$ to $\mathbf{x}$ highlights the high reversibility of our model. Additionally, the test set results suggest strong generalization performance.  

\subsection{Semantic Manipulation}
The semantic manipulation approach can be summarized as finding the path in the latent space that connects the two extremes of the same attribute \cite{wang2019visual}. In practice, the dataset is split into two subsets according to the value of pre-specified attribute. Then, the two subsets are mapped into the latent space and their corresponding average latent vectors can be calculated: $\mathbf{z}_{\mathrm{+}}$ and $\mathbf{z}_{\mathrm{-}}$. The resulting manipulation of a target image $\mathbf{x}_{\mathrm{target}}$ is obtained by adding a scaled manipulation vector $\mathbf{z}_{\mathrm{manipulation}}$ to its corresponding latent variable $\mathbf{z}_{\mathrm{target}}$:
\begin{equation}
\label{eq:semantic}
    \mathbf{z}_{\mathrm{final}} = \mathbf{z}_{\mathrm{target}} + \alpha \cdot \mathbf{z}_{\mathrm{manipulation}},
\end{equation}
where $\mathbf{z}_{\mathrm{manipulation}} = \mathbf{z}_{\mathrm{+}} - \mathbf{z}_{\mathrm{-}}$ and $\alpha$ is the hyperparameter to control the manipulation.
	
\begin{figure}[ht]
    \centering
    \includegraphics[width=0.99\linewidth]{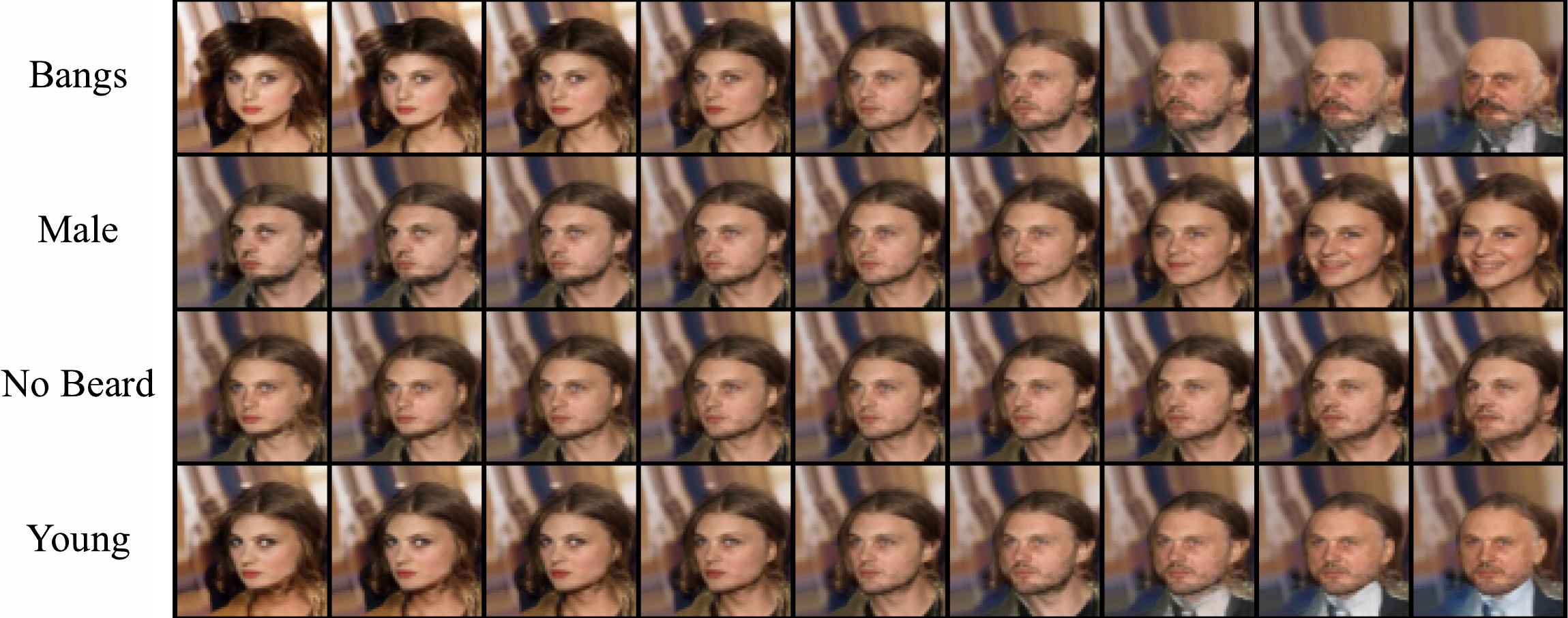}
    \caption{Semantic manipulation of attributes of a face image. Each row interpolates the target image’s latent code along the attribute vector, with the middle image as the original.
    }
    \label{fig:semantic-manipulation}
\end{figure}

We conduct experiments on the CelebA dataset, with results shown in \cref{fig:semantic-manipulation}. Each row is generated by interpolating the latent code of the target image along the attribute vector, with the middle image as the original. The interpolation factor $\alpha$ varies uniformly from -2 to 2 across each row. The results demonstrate that EIW-Flow enables smooth and meaningful facial feature transformations.  
	
\subsection{Interpolation Experiments}
\subsubsection{Comparison of Interpolation}
\begin{figure}[ht]%
    \centering
    \subfloat[CelebA ($64\times64$ pixels).]{\label{fig:celeba-test-recon-inter}\includegraphics[width=0.9\linewidth]{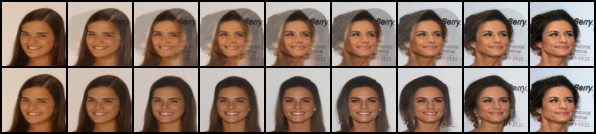}} \\
    \subfloat[CIFAR10 ($32\times32$ pixels).]{\label{fig:cifar10-test-recon-inter}\includegraphics[width=0.9\linewidth]{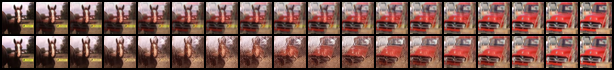}}
    \caption{Comparison of interpolation on CelebA and CIFAR-10 datasets. The first row shows data-space interpolation results  $\mathbf{x}_{\alpha}$, and the second row shows images reconstructed from latent-space interpolation $\mathbf{z}_{\alpha}$.}
    \label{fig:celeba-recon-inter}
\end{figure}

We compare interpolation in data and latent spaces. Two samples $\mathbf{x}_0$ and $\mathbf{x}_1$ are drawn from the dataset and interpolated to obtain  $\mathbf{x}_{\alpha}$ as in \cref{sec:two-points-inter}. Their latent codes $\mathbf{z}_0$ and $\mathbf{z}_1$ are then interpolated to get  $\mathbf{z}_{\alpha}$, which is mapped back to data space and compared with $\mathbf{x}_{\alpha}$. Experiments are conducted on CelebA and CIFAR-10, with results shown in \cref{fig:celeba-test-recon-inter} and \cref{fig:cifar10-test-recon-inter}.

It can be seen that, in both CelebA and CIFAR-10  datasets, interpolation in the latent space tends to produce realistic images in the data space, whereas interpolation in the data space alone may cause unnatural effects (e.g., phantoms, overlaps, etc.). More importantly, this results demonstrates that our model learns a meaningful transformation between the latent space and the data space. In  \cref{fig:celeba-test-recon-inter},  the model learns a transformation that manipulates the orientation of the head. While in \cref{fig:cifar10-test-recon-inter} , the model has the ability to gradually turn a brown horse  into a vibrant red car. It demonstrates that the latent space we have explored can be utilized in tasks that require more local face features such as  3D point clouds generation \cite{li2023progressive}.

\subsubsection{Two-points Interpolation}
\label{sec:two-points-inter}
\begin{figure}[!ht]%
    \centering
    \subfloat[CelebA ($64\times64$ pixels).]{\label{fig:celeba-twopoints-multiple}\includegraphics[width=0.99\linewidth]{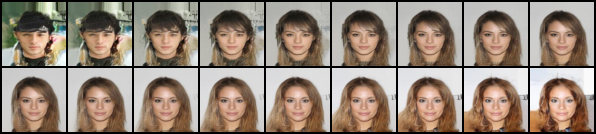}}\\
    \subfloat[CIFAR-10 ($32\times32$ pixels).]{\label{fig:cifar10-twopoint-multiple}\includegraphics[width=0.99\linewidth]{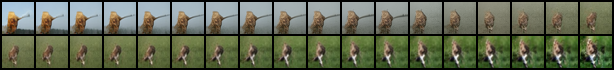}}%
    \caption{Results of two-points interpolation on CelebA and CIFAR-10 datasets. The top left and bottom right images are endpoints $\mathbf{z}_0$ and $\mathbf{z}_1$. The other images corresponds to the middle points $\mathbf{z}_{\alpha} (0<\alpha<1)$. $\alpha$ increases gradually from the left to right and top to bottom.}
    \label{fig:dataset-twopoints-multiple}
\end{figure}

We first randomly sample two datapoints $\mathbf{z}_0$ and $\mathbf{z}_1$ from the last scale. We then generate middle points $\mathbf{z}_{\alpha}$ by $\mathbf{z}_{\alpha}=(1-\alpha)\mathbf{z}_1+\alpha\mathbf{z}_0$. We back-propagate these variables together and interpolate at each scale using the same interpolation method. The CelebA and CIFAR-10 images are shown in \cref{fig:celeba-twopoints-multiple} and \cref{fig:cifar10-twopoint-multiple}. We observe smooth interpolation between images belonging to distinct classes. This shows that the latent space we have explored can be utilized for downstream tasks like image editing. 

\section{Ablation study}
\subsection{Effect of Temperature}
\begin{figure}[!ht]
    \centering
    \includegraphics[width=0.99\linewidth]{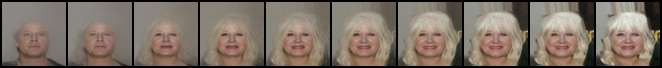}
    \caption{Effect of temperature annealing on CelebA dataset. The images from left to right show the uniform increase of annealing parameter $T$ from $0.5$ to $1$. When $T=1$, no temperature annealing is applied.}
    \label{fig:celeba-temperature}
\end{figure}

We evaluate temperature annealing by scaling latent codes as $T\mathbf{z}$ with $T \in [0,1]$. When $T=1$, no temperature annealing is applied.
The scaled codes are mapped back to data space to generate images (\cref{fig:celeba-temperature}). It can be seen that taking value of annealing parameter $T$ from $0.8$ to $1$, i.e., four images on the right, can obtain realistic images and preserve face details from the original image. However, when $T<0.8$, the background and hair gradually becomes void as $T$ decreases.

\subsection{Effect of $\mathtt{Shuffle}$ operation}
\label{sec:effectiveness-of-shuffle}
To demonstrate the effectiveness and necessity of the $\mathtt{Shuffle}$ operation, we  conduct an ablation experiment to compare the channel feature maps of latent variables before and after the $\mathtt{Shuffle}$ operation.

In \cref{multiscale}, we mention that $\mathbf{x}_k$ is propagated to the next scale, while $\mathbf{z}_k$ remains unchanged to form $\mathbf{z}$. However, the visual difference between the channel feature maps of $\mathbf{x}_k$ and $\mathbf{z}_k$, as well as the difference before and after the $\mathtt{Shuffle}$ operation, has not been evaluated. To address these, we conduct an ablation experiment following the procedure outlined below. For simplicity, we continue using $\mathbf{x}_k$ and $\mathbf{z}_k$ to refer to the first and second halves of the latent variables $\mathbf{u}_k$ and $\hat{\mathbf{u}}_k$, regardless of whether it's before or after the $\mathtt{Shuffle}$ operation.

\begin{figure*}[!ht]
    \centering
    \includegraphics[width=0.99\linewidth]{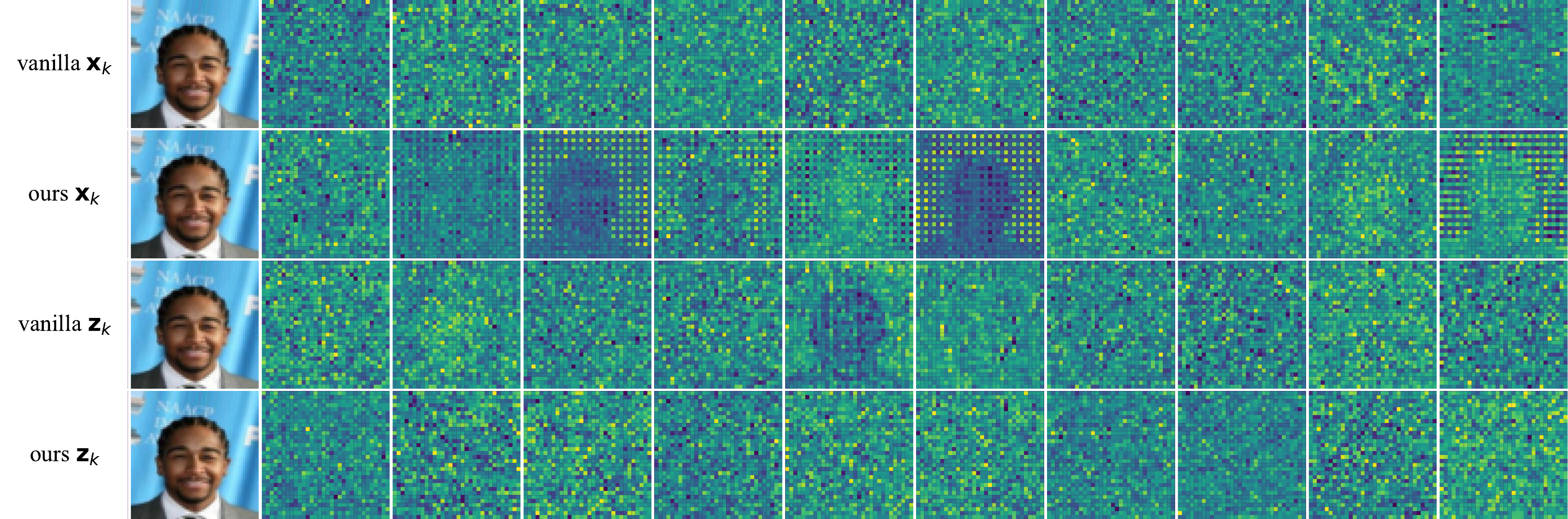}
    \caption{Effectiveness of $\mathtt{Shuffle}$ operation on CelebA dataset. The first and third rows represent the channel feature maps before the $\mathtt{Shuffle}$ operation, and the other two rows represent the channel feature maps after the $\mathtt{Shuffle}$ operation. The terms $\mathbf{x}_k$ and $\mathbf{z}_k$ indicate the first and second halves of both latent variables $\mathbf{u}_k$ and $\hat{\mathbf{u}}_k$, see  \cref{sec:effectiveness-of-shuffle} for details.}
    \label{fig:celeba-middle-status-train}
\end{figure*}

Both before and after the $\mathtt{Shuffle}$ operation, we choose $l<C_k$ channel feature maps of $\mathbf{x}_k$ and $\mathbf{z}_k$, whose corresponding values in $\mathbf{Q}_{\boldsymbol{\phi}}$ are greater than others. Similarly, we can also choose $l$ channels with lower values in $\mathbf{Q}_{\boldsymbol{\phi}}$. In this experiment, we set $l = 10$ for a better visualization.  Results are shown in  \cref{fig:celeba-middle-status-train}. In this figure, the terms "vanilla" and "ours" are used to refer to the results before and after the $\mathtt{Shuffle}$ operation, respectively. These two terms also corresponds to the results without and with the $\mathtt{Shuffle}$ operation, respectively.

We can observe that the channel feature maps of $\mathbf{z}_k$ before the $\mathtt{Shuffle}$ operation (i.e. vanilla $\mathbf{z}_k$) still contain some face structure information, while corresponding  channel feature maps of $\mathbf{x}_k$ (i.e. vanilla $\mathbf{x}_k$) seem more like Gaussian noise. This result contradicts with the purpose of a multi-scale architecture, i.e. $\mathbf{z}_k$ ought to follow a multivariate Gaussian distribution and $\mathbf{x}_k$ should contain some feature information. However, the results after the $\mathtt{Shuffle}$ operation are contrary to the previous ones. This implies the channel feature maps of $\mathbf{x}_k$ contain more face structure information and $\mathbf{z}_k$ are more likely to follow a multivariate Gaussian distribution. This provides evidence to demonstrate the effectiveness of our designed $\mathtt{Shuffle}$ operation.

\subsection{Effect of hyperparameter $\lambda$}

In \cref{eq:final-loss}, we introduce a hyperparameter, $\lambda$, to balance between the vanilla and our introduced loss. If $\lambda$ is too small, the \emph{solver}-$\mathcal{S}$ may fail to approximate \emph{guider}-$\mathcal{G}$, and when $\lambda=0$, $\mathcal{G}$ stops working, causing disorganized shuffling of latent variables by the $\mathtt{Shuffle}$ operation. On the other hand, if $\lambda$ is too large, the likelihood loss function's effect on EIW-Flow is diminished, reducing its expressive power. Thus, selecting an appropriate $\lambda$ is crucial for training EIW-Flow. To find the optimal value, we conduct an ablation experiment on the CIFAR-10 dataset with NFs like EIW-Flow.  

We apply the $\mathtt{Shuffle}$ operation to CNFs like TO-FLOW \cite{du2022flow} on the CIFAR-10 dataset. This extended ablation experiment examines the effect of $\lambda$ across different NF types. The results are shown in \cref{tab:lambda}. 
For CNFs, the results were obtained after training for $50$ epochs, which took about seven days on a single Titan V GPU. For NFs, $\lambda=1e-3$ yields the best performance $2.97$, with performance degrading for larger or smaller values of $\lambda$. For CNFs, $\lambda=1e-4$ provides the best performance.
\begin{table}[ht]
    \caption{Ablation study on hyperparameter $\lambda$ on CIFAR-10 dataset for various types of NFs, i.e., NFs and CNFs. }
    
    \centering
    \begin{tabular}{cc|cc}
        \hline
        \multicolumn{2}{c|}{\textbf{Normalizing Flows (NFs)}} & \multicolumn{2}{c}{\textbf{Continuous Normalizing Flows (CNFs}} \\
        \hline
        hyperparameter $\lambda$ & \textbf{bits/dim}$\downarrow $ & hyperparameter $\lambda$ & \textbf{bits/dim}$\downarrow $\\
        \hline
        $1e-1$ & $3.092$ & $1e-1$ & $3.5025$ \\
        $1e-2$ & $3.021$ & $1e-2$ & $3.5097$ \\
        $1e-3$ & $2.97$ & $1e-3$ & $3.5024$ \\
        $1e-4$ & $3.015$ & $1e-4$ & $3.4981$ \\
        \hline
    \end{tabular}
    \label{tab:lambda}
\end{table}
	
\section{Limitations and Future Work}
Our work follows standard normalizing flow conventions, using Gaussian target distributions for theoretical guarantees via the maximum entropy principle while achieving strong empirical results.
Future extensions include  computational optimizations and theoretical adaptations for specialized domains.
First, extending EIW-Flow to non-Gaussian or learned priors could further enhance model flexibility, while evaluation on non-curated datasets would improve robustness and generalization. In addition, although our shuffle mechanism introduces negligible overhead (Sec. 4.5), future work could explore further acceleration strategies and hardware-aware designs to benefit large-scale training. Beyond flow-based models, integrating our transformation module into diffusion-based frameworks could reduce the number of denoising steps and speed up the noise-to-data process. From an application perspective, EIW-Flow is particularly well-suited for tasks requiring both expressive density estimation and fine-grained feature preservation, such as conditional image rescaling \cite{zha2023conditional}, optimal experimental design \cite{dong2025variational} and knowledge graph reasoning \cite{ma2025historical}. Finally, inspired by CAST \cite{cao2024cast}, LSTNet \cite{ma2023towards}, and DTNet \cite{ma2024image}, future work could explore integrating entropy-guided feature routing into vision-language tasks such as multimodal captioning or retrieval to further improve cross-modal alignment. Collectively, these directions highlight EIW-Flow’s potential to impact a wide range of real-world applications while maintaining mathematical elegance and practical scalability.

\section{Conclusion}
In this paper, we propose a reversible and regularized $\mathtt{Shuffle}$ operation and integrate it into vanilla multi-scale architecture. The resulting flow-based generative model is called Entropy-informed Weighting Channel Normalizing Flow (EIW-Flow). 
The $\mathtt{Shuffle}$ operation is composed of three distinct components: \emph{solver}, \emph{guider} and \emph{shuffler}. We demonstrate the efficacy of the $\mathtt{Shuffle}$ operation from the perspective of entropy using Central Limit Theorem and the Maximum Entropy Principle. 
The quantitative and qualitative experiments show that EIW-Flow results in better density estimation and generates high-quality images compared with previous state-of-the-art deep generative models. 
Besides, we compare the computational complexity with the vanilla architecture and observe that only negligible additional computational overhead is added by our model.
Furthermore, we also conduct different types of ablation experiments to explore the effect of hyperparameters and the  $\mathtt{Shuffle}$ operation.
	
\section*{Acknowledgement}
The work is supported by the Fundamental Research Program of Guangdong, China, under Grants 2020B1515310023
and 2023A1515011281; and in part by the National Natural Science Foundation of China under Grant 61571005.

\bibliographystyle{elsarticle-num}
\bibliography{egbib.bib}

\end{document}